\documentclass[10pt]{article} 
\usepackage[preprint]{tmlr}

\usepackage{amsmath,amsfonts,bm}

\def\eqref#1{equation~\ref{#1}}

\def\1{\bm{1}}

\DeclareMathAlphabet{\mathsfit}{\encodingdefault}{\sfdefault}{m}{sl}
\SetMathAlphabet{\mathsfit}{bold}{\encodingdefault}{\sfdefault}{bx}{n}

\newcommand{\R}{\mathbb{R}}

\usepackage[dvipsnames]{xcolor}
\usepackage{hyperref}
\usepackage{url}
\usepackage{algorithm}
\usepackage{algpseudocode}
\usepackage{colortbl}
\usepackage{booktabs}       
\usepackage{multirow}
\usepackage{amssymb}
\usepackage{amsthm}
\usepackage{amsfonts}
\usepackage{wrapfig}
\usepackage{xfrac}
\usepackage{pgfplots}
\usepgfplotslibrary{groupplots}
\usetikzlibrary{patterns}
\pgfplotsset{compat=newest}
\usepackage{algorithm}
\usepackage{algpseudocode}
\definecolor{LightCyan}{rgb}{0.75,1,1}
\makeatletter
\pgfplotsset{
    groupplot xlabel/.initial={},
    every groupplot x label/.style={
        at={($({\pgfplots@group@name\space c1r\pgfplots@group@rows.west}|-{\pgfplots@group@name\space c1r\pgfplots@group@rows.outer south})!0.5!({\pgfplots@group@name\space c\pgfplots@group@columns r\pgfplots@group@rows.east}|-{\pgfplots@group@name\space c\pgfplots@group@columns r\pgfplots@group@rows.outer south})$)},
        anchor=north,
    },
    groupplot ylabel/.initial={},
    every groupplot y label/.style={
            rotate=90,
        at={($({\pgfplots@group@name\space c1r1.north}-|{\pgfplots@group@name\space c1r1.outer
west})!0.5!({\pgfplots@group@name\space c1r\pgfplots@group@rows.south}-|{\pgfplots@group@name\space c1r\pgfplots@group@rows.outer west})$)},
        anchor=south
    },
    execute at end groupplot/.code={%
      \node [/pgfplots/every groupplot x label]
{\pgfkeysvalueof{/pgfplots/groupplot xlabel}};  
      \node [/pgfplots/every groupplot y label] 
{\pgfkeysvalueof{/pgfplots/groupplot ylabel}};  
    }
}

\def\endpgfplots@environment@groupplot{%
    \endpgfplots@environment@opt%
    \pgfkeys{/pgfplots/execute at end groupplot}%
    \endgroup%
}
\makeatother

\definecolor{darkblue}{rgb}{0, 0, 0.5}
\hypersetup{colorlinks=true, citecolor=darkblue, linkcolor=darkblue, urlcolor=darkblue}

\newenvironment{usethmcounterof}[1]{%
  \renewcommand\thethm{\ref{#1}}\thm}{\endthm\addtocounter{thm}{-1}}
\newcommand{\method}{{{Compactor}}\,}
\newcommand{\LM}{\textsf{LM}\,}
\newcommand{\K}{\mathbf{K}}

\newtheorem{defi}{Definition}
\newtheorem{cor}{Corollary}
\newtheorem{thm}{Theorem}

\title{Compactor: Calibrated KV Cache Compression with \\Approximate Leverage Scores}

\author{\name Vivek Chari \email vchari2@jhu.edu \\
      \addr Department of Computer Science\\
      Johns Hopkins University
      \AND
      \name Benjamin Van Durme \email vandurme@jhu.edu \\
      \addr Department of Computer Science\\
      Johns Hopkins University}

\def\month{MM}  
\def\year{YYYY} 
\def\openreview{\url{https://openreview.net/forum?id=XXXX}} 

\begin{document}

\maketitle

\begin{abstract}
Modern Large Language Models (LLMs) are increasingly trained to support very large context windows. Unfortunately the ability to use long contexts in generation is complicated by the large memory requirement of the KV cache, which scales linearly with the context length. This memory footprint is often the dominant resource bottleneck in real-world deployments, limiting throughput and increasing serving costs. One way to address this is by compressing the KV cache, which can be done either with knowledge of the question being asked (query-aware) or without knowledge of the query  (query-agnostic). 
We present \method, a training-free, query-agnostic KV compression strategy that uses approximate leverage scores to determine token importance. We show that \method can  achieve the same performance as competing methods while retaining 20\% fewer tokens in both synthetic and real-world context tasks, while being far more task-robust. We further introduce a procedure for \emph{context-calibrated compression}: inferring the maximum compression a given context supports before significant performance loss. Using context-calibrated compression, we show that \method achieves full KV performance on Longbench while reducing the KV memory burden by 68\%, on average. To demonstrate the efficacy and generalizability of our approach, we apply \method to 27 synthetic and real-world tasks from RULER and Longbench, with models from both the \texttt{Qwen 2.5} and \texttt{Llama 3.1} families. Finally, we release \texttt{compactor-vllm}\footnote{Our code can be found at \url{https://github.com/vnchari/compactor-vllm}.}, an inference engine and suite of optimized Triton kernels designed to efficiently support the sparse, non-contiguous memory access patterns inherent to compressed KV caches. This work demonstrates that \method offers a practical, high-performance solution for alleviating the memory bottleneck in modern LLM deployment.
\end{abstract}
\section{Introduction}
Much effort has been devoted to increasing the context length supported by modern Large Language Models (LLMs). While many LLMs now support incredibly long context windows, practical use of long documents is complicated by the autoregressive attention mechanism: at each generation step we must retain all key/value pairs for every past token, so the KV cache grows linearly with context length. For example, running \texttt{Qwen-2.5 32B} on a prompt of size 100K requires maintenance of 26 GB of KV cache, per request; such a memory burden limits the scalability of long-context deployment. This memory burden is especially problematatic for inference servers, given the widespread use of prompt caching.

To address this, two orthogonal lines of work have emerged. One has focused on improving LLM generation efficiency by careful management of the KV cache  \citep{kwon_efficient_2023,zheng_sglang_2024}, and designing more complex prompt caching strategies \citep{gim_prompt_2024}. The other has focused on methods that reduce the memory burden of the KV cache by compressing, evicting, or summarizing tokens. Less work has focused on the integration of the aforementioned techniques.

We identify two key barriers to the integration of KV compression techniques into efficient LLM serving systems. Firstly, most compression schemes rely on knowing the query at compression time; in this query-aware setting, the query is used to inform which tokens are evicted. Unfortunately, the performance of query-aware compression methods degrades significantly when the query is not known at compression time (the query-agnostic regime) \citep{chari_kv-distill_2025,li_scbench_2025}. Additionally, query‐aware schemes are incompatible with multi‐query prefix caching: each new question would evict different prefix tokens, so the compressed prefix cannot be reused across prompts. Second, even when query-agnostic compression is possible, determining how many tokens can be evicted without degrading quality is non-trivial. Such determinations are especially needed in applications that lack a prescribed token or memory budget, but where generation quality is paramount. To the best of our knowledge, there has been little prior work targeting this issue of  \emph{context-calibrated compression}.

\paragraph{Contributions} 
In this work we present \method, a lightweight, (trainable) training-free KV-eviction strategy that operates without query knowledge and still preserves answer quality. Our contributions are as follows:
\begin{enumerate}
    \item We propose \method, a two-component eviction mechanism that uses leverage scores and non-causal attention scores to significantly outperform current query-agnostic mechanisms. We further propose a fast randomized algorithm to approximate the aforementioned leverage scores to provide a fast, practical implementation of \method, and prove guarantees regarding the quality of the output of the randomized algorithm.
    \item We introduce the notion of \emph{context-calibrated compression}, which allows one to infer the maximum compression a particular context can support with minimal inference-time overhead. Our formulation of context-calibration can be applied to any choice of eviction policy. Figure~\ref{fig:longbench_calibrated_bars} shows that \method can achieve significantly higher query-agnostic compression rates than competing methods. 
    \item We evaluate \method across multiple LLMs, and both synthetic and real-world long-context datasets. We show that on Longbench, \method performs comparably to using the full KV cache, while retaining a third of the KV cache. On synthetic UUID needle retrieval tasks, we show that we can achieve full performance while retaining only 50\% of the cache. 
    \item We release \texttt{compactor-vllm}, an inference engine with a custom paged KV cache implementation and a suite of custom Triton kernels to make KV cache compression practically feasible. 
\end{enumerate}
\section{Background}

We focus on a Transformer based language model \citep{vaswani_attention_2023}. Applying the standard self‑attention mechanism to a context of length $N$ (“prefilling’’) has time complexity $O(N^{2})$, while auto‑regressively decoding a single token costs $O(N)$; storing the resulting key–value (KV) cache requires $O(N)$ memory.  Prefilling is therefore usually compute‑bound, whereas decoding is dominated by memory bandwidth.  A large body of work tackles these bottlenecks by manipulating the KV cache at different points in the generation pipeline.  To describe this work, we adopt the taxonomy of \citet{li_scbench_2025}, which divides KV cache management into 4 stages: \emph{KV generation}, \emph{KV retrieval}, \emph{KV loading}, and \emph{KV compression}. 

\noindent\textbf{KV Generation} \; The aim at this stage is to construct the cache more cheaply during prefilling. Recent work pursues attempts to sparsify attention patterns during prefilling\citep{beltagy_longformer_2020,xiao_efficient_2024,jiang_minference_2024}; these works largely differ in the particular sparsity mask they employ.  Another line of work tries to replace self‑attention altogether with recurrent models that have bounded state sizes \citep{gu_mamba_2024,lieber_jamba_2024}. Yet another line of work focuses on cache-aware scheduling during the prefilling stage \citep{fu_lazyllm_2024, zhao_alisa_2024,cho_kv-runahead_2024}. Finally, prompt‑level compression can be applied \emph{before} the model to drop unimportant tokens, lowering prefill cost proportionally \citep{pan_llmlingua-2_2024}. Such architectural changes are usually query-agnostic.

\noindent\textbf{KV Retrieval \& Loading} \;Once a cache exists, the system must decide \emph{which} cached segments to reuse across requests. Efficient inference frameworks already exploit exact prefix matching\citep{kwon_efficient_2023,zheng_sglang_2024}. When the desired KVs reside in slower memory, they must be loaded to GPU efficiently. For example, PagedAttention in vLLM treats GPU RAM as a pool of fixed‑size pages, enabling fine‑grained sharing of KV segments across requests. Similarly, \citep{qin_mooncake_2024} separate the inference workload into prefilling and a decoding cluster, and use a disaggregated KV store to improve (CPU) RAM and SSD utilization. Beyond exact reuse, \emph{semantic} retrieval matches new prompts to a vector database of cached prefixes, by either skipping the prefill when the prefix similarity exceeds a learned threshold \citep{yao_cacheblend_2024}, or anticipates which KV elements will be attended to and pre-fetching them \citep{lee_infinigen_2024}. Such methods are inherently query-aware.

\noindent\textbf{KV Compression} \; The final line of defense is to reduce the footprint of the cache. Quantization is the most direct route, and reduces the memory footprint of each element in the KV cache \citep{liu_kivi_2023,zhang_kv_2024}.
Orthogonal works prune tokens altogether to reduce the sequence dimension. StreamingLLM keeps only a sliding window of recent tokens \citep{xiao_efficient_2024}; $\text {H}_2\text{O}$, SnapKV, and PyramidKV use accumulated attention scores to drop the least‑attended tokens (see \S \ref{sec:attn_based-scoring} for further details) \citep{zhang_h_2o_2023, li_snapkv_2024,cai_pyramidkv_2024}. Variants of the above methods apply different compression ratios or strategies to different heads \citep{tang_razorattention_2024, feng_ada-kv_2025}. Token-dropping techniques work well in the query-aware setting, because the attention that tokens in the question place on tokens in the context are used to select necessary tokens. However, in the query-agnostic regime, their performance suffers greatly \citep{li_scbench_2025, chari_kv-distill_2025}.

Each of these techniques improves KV‑cache efficiency along an orthogonal dimension and can therefore be composed to yield highly efficient inference procedures. In this work we focus on KV‑cache compression along the sequence dimension; accordingly, we compare only against methods that explicitly reduce the KV cache size via token-dropping in our experiments. 

\section{\method}
\noindent\textbf{Preliminaries \& Overview} \; Let a sequence consist of $N$ tokens, each mapped to $d$‑dimensional key and value vectors collected in matrices $\K, \mathbf V\in\mathbb R^{N\times d}$.  Standard self‑attention caches the entire $(\K, \mathbf V)$ matrices, incurring $O(Nd)$ memory burden.  We aim to retain only a fraction $r\in(0,1]$ of the tokens while preserving a model's predictive quality.  While not an asymptotic improvement in the memory usage, in practice one can achieve significant memory savings with minimal performance loss by utilizing token eviction. To determine which tokens are to be evicted/discarded, we compute a vector $\vec s \in \mathbb R^N$ of ``importance scores'' and select the top-$k$ with $k=\lceil r\cdot N\rceil$ highest scoring tokens for retention. To compute $\vec s$, we introduce two complementary score families: geometry-based outlier scores $\vec o$ and task-driven attention scores $\vec a$; each captures different, essential, aspects of the key distribution. We blend them to generate final importance scores $\vec s$.

\noindent\textbf{Notation} \;We consider a transformer-based language model \citep{vaswani_attention_2023}, denoted by \LM, over a vocabulary $\mathcal V$. Let $c \in \mathcal V^N$ denote a context, and with slight abuse of notation, let $\tilde c_r$ denote (the KV cache of) a context that has been compressed with retention rate $r$ (i.e if $r=0.1$, we retain 10\% of tokens). Throughout, indices $i,j\in[N]$ refer to tokens, whereas boldface capitals denote matrices.  For a matrix $\mathbf A\in\mathbb R^{m\times n}$, $A_i$ is its $i$‑th row, $\operatorname{rank}(\mathbf A)$ its rank, and $\mathbf A^\top$ its transpose. We denote the total, outlier, and attention ``importance score'' vectors as $\vec s, \vec o, \vec a$ respectively. Let $\bar{{v}}$ denote the average value of a vector $\vec{{v}}$, and $\operatorname{std}(\vec{{v}})$ denote the standard deviation of the values of the vector. For clarity, we do not notate the head dimension, as the same procedure is applied to heads independently. 

\subsection{Outlier Scores via Statistical Leverage}
We first turn our attention to generating ``outlier'' scores $\vec o \in \R^N$. Retaining merely the most‐attended keys fails when a key is rarely queried yet encodes unique information, i.e., it is an \emph{outlier} in the key space. Eviction of these \emph{outliers} leads to irreversible information loss. We formalize the notion of ``outlierness'' via the classical concept of statistical leverage:

\begin{defi}[Leverage Score]
    Let $\K \in \R^{N \times d}$ and denote its SVD as ${\textsc{svd}(\K) = \mathbf U \mathbf \Sigma \mathbf V^\top}$. Then the leverage score of the $i$-th row of $K$
    \[
    \ell_i = ||U_i||_2^2  \qquad \qquad \sum_{i=1}^{N} \ell_i = \operatorname{rank}(k)
    \]
\end{defi}
Intuitively, rows whose associated left‐singular vectors have large Euclidean norm align closely with the directions in which $\K$ has the greatest variance. Omitting these rows would discard significant information. Theorem~\ref{thm:leverage_scores} formalizes the  utility of leverage scores in row selection. 
\begin{thm}[Spectral Preservation of Leverage Sampling]
\label{thm:leverage_scores}
    Let $\epsilon, \delta \in (0, 1)$ and data matrix $\K \in \R^{N \times d}$ be given and take $k\geq Cd\log(\frac{d}{\delta})\epsilon^{-2}$ for some universal constant $C$. Construct $\hat{\K}_k \in \R^{k \times d}$ by sampling $k$ times (with replacement) from the distribution that places probability mass $\tfrac{\ell_i}{\operatorname{rank}(\K)}$ on the $i$-th row. Scale each of the $k$ chosen rows by $\sqrt{d/(k\ell_i)}$. Then, with probability $1 - \delta$, we have 
    \[
    (1 - \epsilon) \K^\top \K \preccurlyeq \hat{\K}^\top_k \hat{\K}_k \preccurlyeq (1 + \epsilon) \K^\top \K
    \]
    where $A \preccurlyeq B$ means that $B - A$ is PSD. In practice \citep{broadbent_subset_2010,paschou_pca-correlated_2007, jolliffe_discarding_1972}, even deterministically choosing rows corresponding to the top-$k$ leverage scores provides a good spectral sparsification of $K$.
\end{thm}
The above theorem provides an intuitive explanation of why leverage scores are useful in choosing which tokens in the KV cache to retain. As such, we define outlier scores as follows: 
\begin{equation}
    \mathbf K =  \mathbf {U\Sigma V}^\top \qquad\qquad \vec o  = \operatorname{diag}(\mathbf U^\top\mathbf U) = \begin{bmatrix} \ell_1 & \ell_2 & \dots & \ell_n\end{bmatrix}
\end{equation}
where $\mathbf K =  \mathbf {U\Sigma V}^\top$ denotes the SVD of pre-positionally encoded keys. Naively computing leverage scores by computing the SVD of an $n \times d$ matrix $\mathbf K = \mathbf{U \Sigma V}^\top$ is much too slow in practice. Instead we utilize the following simple identity: 
\begin{equation}
\label{eq:fast_svd}
    \textsc{svd}(\K^\top \K) = \mathbf V \mathbf \Sigma^2\mathbf V^\top \implies \mathbf U = \K \mathbf V \mathbf \Sigma^{-1}
\end{equation}
so we can instead construct $\mathbf U$ by computing the SVD of a $d \times d$ matrix and performing two GEMMs. Though this procedure is asymptotically the same complexity, hardware-level optimization of GEMM routines makes Equation~\ref{eq:fast_svd} significantly faster when $d \ll N$ (as is the case in the KV cache). In the next section, we further reduce the time to compute the leverage scores by approximating them. 
\subsection{Approximate Leverage Score Computation}

Given the computational intensity of computing leverage scores, there has been substantial work in developing methods for approximating them \citep{drineas_fast_2012, chen_fast_2021,eshragh_salsa_2023,zuo_quantum-inspired_2021}. A key technique in most leverage score approximation algorithms is ``sketching'' the original matrix to obtain a smaller matrix that roughly preserves the properties of the original, in the following sense: 
\begin{defi}[Subspace Embedding]
     Let $L \subseteq \mathbb R^d$ be a linear subspace. Let $\mathbf \Phi: \R^d \to \R^k$ be a linear map with the property that $\forall x \in L$
     \[
     (1 - \epsilon)^2 ||x||_2^2 \leq ||\mathbf\Phi x||_2^2\leq (1 + \epsilon)^2 ||x||_2^2 
     \]
     The map $\mathbf\Phi$ is called a subspace embedding (or "sketch") for $L$ with embedding dimension $k \leq d$ and distortion $\epsilon > 0$.
\end{defi}

We refer to $\mathbf\Phi$ as a ``sketching" matrix. In contrast to prior work in leverage-score approximation, which applies left sketches ($\mathbf\Phi \K$), we apply a right sketch ($\K \mathbf \Phi$) to reduce the dimensionality of the column-space of the data, at the cost of weaker theoretical bounds on the error in the approximate leverage scores. The reason for using right sketching is three-fold: (1) speeding up computation of Eq~\ref{eq:fast_svd} requires a right sketch; (2) instantiating a left sketching matrix and subsequently computing $\mathbf \Phi \mathbf K$ requires substantial memory when $N$ is large; and (3) in practice, we observe no degradation in the utility of leverage scores for KV cache approximation when right sketching (see \S~\ref{sec:ablation}). The algorithm to compute approximate leverage scores is presented in Algorithm~\ref{alg:leverage-sampling}.  
    \begin{algorithm}[H]
  \caption{Approximate Leverage Scores}
  \label{alg:leverage-sampling}
  \begin{algorithmic}[1]
    \Require $\K\in\R^{N\times d}$, target dimension $k$
    \Ensure Approximate leverage Scores $\vec o$
    \State $\mathbf\Phi \in \R^{d \times k}$ with $\Phi_{ij} \sim \mathcal N(0, \frac{1}{k})$
    \State $\hat {\K} \gets \mathbf{K \Phi}$
    \State $\mathbf G \gets \hat {\K}^\top \hat {\K}$
    \State $\textsc{svd}(\mathbf G) = \tilde {\mathbf V} {\mathbf \Sigma}^2 \tilde{\mathbf V}^\top$ \Comment{SVD of a $k \times k$ matrix}
    \State $\tilde {\mathbf U} \gets \hat {\K} \tilde {\mathbf V} \tilde {\mathbf \Sigma}^{-1}$
    \State \Return $\vec o \gets \operatorname{diag}(\tilde {\mathbf U} \tilde{\mathbf{U}}^\top)$
  \end{algorithmic}
\end{algorithm}
The quality of the (Gaussian) right-sketching leverage score approximation is formalized in the following theorem: 
\begin{thm}[Approximate Leverage Scores]
\label{thm:approx_lev_scores}
    Let data matrix $\K \in \R^{N \times d}$ and target dimension $k$ be given. Let distortion factor $\epsilon \in (0, 1)$ and failure probability $\delta \in (0, 1)$ be given. Define $\mathbf \Phi \in \mathbb R^{d \times k}$ be a matrix whose entries are drawn i.i.d from $\mathcal N (0, \frac{1}{k})$. Compute the SVD of the sketched matrix $\mathbf {K \Phi}$ and approximate leverage scores $\tilde \ell_i$ as: 
    \[
    \textsc{svd}\left(\K\mathbf\Phi \right)= \mathbf {\tilde U} \mathbf{\tilde \Sigma} \mathbf{\tilde V}^\top \qquad \qquad \tilde \ell_i = ||\tilde U_i||_2^2 
    \]
    If $k \geq 12 \epsilon^{-2} \left(\operatorname{rank}(\K) \log (42\epsilon^{-1}) + \log(2\delta^{-1})\right)$
    Then we have that with probability $1- \delta$
    \[
    \kappa(\K)^{-1}\frac{1 - \epsilon}{(1 + \epsilon)} \ell_i \leq  \tilde \ell_i \leq  \kappa(\K) \frac{1 + \epsilon}{1 - \epsilon} \ell_i \qquad \qquad \forall i \in [N]
    \]
    where $\kappa(\K)$ is the condition number of $\K$.
\end{thm}
The proof of the theorem is deferred to Appendix \ref{appendix:proof_thm_leverage_approximation}. 
Other choices of sketching matrix are also admissible. In Appendix \ref{appendix:srht}, we show that choosing $\Phi$ to be a Subsampled Randomized Hadamard Transform doesn't impact downstream performance, further demonstrating the robustness of the technique. Furthermore, note that the particular construction of $\tilde \ell_i$ shown in Algorithm~\ref{alg:leverage-sampling} is not the only one possible; one can also use an economy QR decomposition, or eigendecomposition. We find these also work well in practice. Appendix C discusses other methods for computing approximate leverage scores, which may be more suitable for other hardware or for highly ill-conditioned key matrices. In summary, we apply Algorithm~\ref{alg:leverage-sampling} to the (pre-position embedded) key-embeddings to generate outlier scores $\vec{o}$. With (approximate) outlier scores in hand, we next derive complementary attention-based scores.

\subsection{Attention-Based Scoring}
\label{sec:attn_based-scoring}
While the leverage scores $\vec o$ capture the geometry of $\K$, attention scores capture the task-specific relevance of each token. We now turn our attention to the latter. Using attention-based scores to inform eviction is not a novel proposition. For example $\textsf{H}_2\textsf{O}$ employs the following scoring mechanism, which computes the accumulated attention (over all queries) for each key: 
\begin{equation}
    \vec{{s}} = {1}^\top \operatorname{softmax}\left( \mathbf Q\mathbf K^\top + \mathbf M\right)
\end{equation}
where ${1}^\top \in \mathbb R^{1 \times n}$ is a vector of all ones and $\mathbf M \in \mathbb R^n$ is an upper triangular matrix filled with $-\infty$ to enforce causality, and selects the top-$k$ highest scoring tokens for retention. Similarly, SnapKV uses the attention of the last $W$ queries (on all keys) to generate  importance scores:
\begin{equation}
    \mathbf Q' = \begin{bmatrix}
        Q_{N - W}^\top &
        Q_{N - W + 1}^\top & 
        \dots & Q_{N}^\top
    \end{bmatrix}^\top  \qquad \vec{{s}} = {1}^\top \operatorname{softmax}\left( \mathbf Q'\mathbf K^\top + \mathbf M'\right)
\end{equation}
where $\mathbf M'$ is a causal mask similar to $\mathbf M$ with appropriate dimensions. The above mechanism is effective because tokens in $W$ can attend to all prior tokens, and especially effective if the window $W$ contains the query to be answered. However, in the query-agnostic setting there is no reason \emph{a priori} to believe that a window of the last $W$ tokens is more informative than any other window. 
\begin{figure}[b]

  \centering
  \scalebox{0.8}{
  \includegraphics[width=0.30\textwidth]{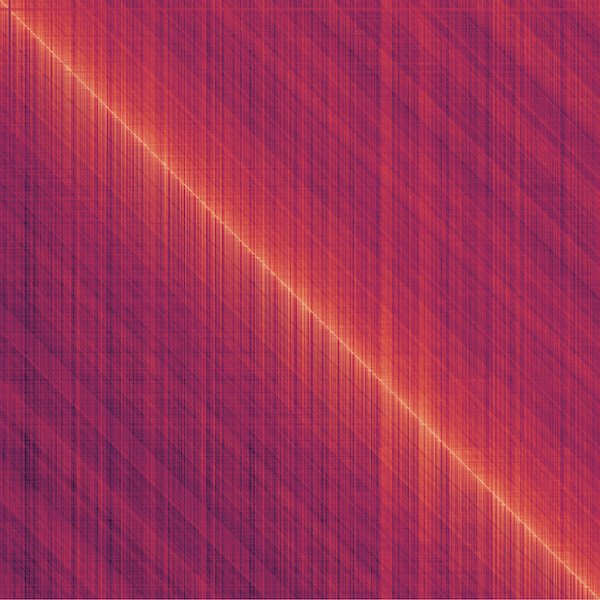}\hspace{0.03\textwidth}
  \includegraphics[width=0.30\textwidth]{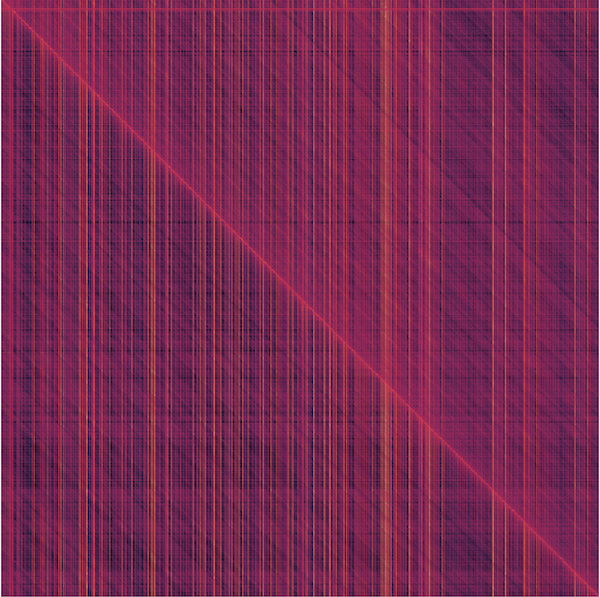}\hspace{0.03\textwidth}
  \includegraphics[width=0.30\textwidth]{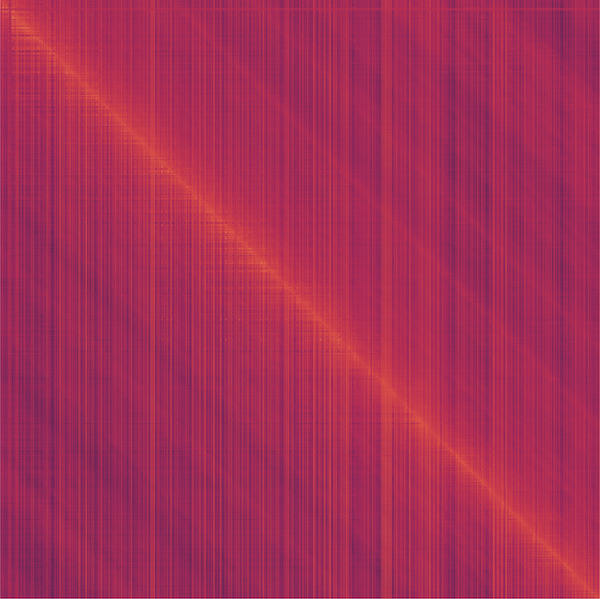}
}
  \caption{ Example of non-causal attention matrices from heads 5, 6, 7 in layer 16 in \texttt{Llama 3.1-8B Instruct}. Lighter colors indicate higher attention scores. Note the columnar and diagonal structure in the non-causal upper-triangular regions of the matrices}
  \label{fig:non_causal_attn_mats}
\end{figure}

Rather than privileging the final $W$ tokens, we desire a query-agnostic statistic that reflects the likelihood that {any} past token will be attended by {any} future one.  Dropping the causal mask and considering the full attention matrix $\mathbf A = \operatorname{softmax}\!\bigl(\mathbf Q\mathbf K^\top\bigr)$ achieves this: each entry $A_{ij}$ measures the model’s tendency to route information from position $j$ to position $i$, independent of generation order. We observe that non-causal attention maps exhibit stable, interpretable features aligned with separator tokens, and few-shot exemplars (see Figure~\ref{fig:non_causal_attn_mats} for a representative example; see Appendix B for an extended discussion of this structure). Accordingly, we define the attention-based importance score as the column-wise sum $\vec{{a}} = \mathbf{1}^\top \operatorname{softmax}\left( \mathbf Q\K^\top\right)$. Note that computation of the above requires instantiation of the entire attention matrix, which is prohibitively expensive. To avoid this, we chunk the input context into blocks of size $C$ and compute scores over each chunk:
\begin{equation}
    \vec{{a}} = \operatorname{concat}_{i=1}^{\lceil N / C \rceil} \left( {1}^\top \operatorname{softmax}\Bigl(\mathbf Q[i]\K[i]^\top\Bigr) \right)
\end{equation}
 where \( \mathbf Q[i] \) (and similarly \( \K[i] \)) is the submatrix containing rows \((i-1)C+1\) through \( iC \). We apply mean-pooling to $\vec{{a}}$ as suggested by \citep{li_snapkv_2024}.
\subsection{Score Blending and KV Selection}
We combine score types into a single ranking signal to inform eviction.
We blend the outlier (leverage) scores $\vec o$ and attention scores $\vec a$ with hyperparameter $\lambda$ to compute final token-scores as 
\begin{equation}\label{eq:scoring}
    \vec s = \frac{\vec{{a}} - \bar{a}}{\operatorname{std}(\vec{{a}})} + \lambda \cdot \left(\frac{\vec{{o}} - \bar{o}}{\operatorname{std}(\vec{{o}})}\right)
\end{equation}
The \method eviction procedure is then: (1) compute token scores $\vec s$ as described by Eq~\ref{eq:scoring}; (2) retain the top-$\lceil r \cdot N\rceil$ tokens with the highest scores (either across all heads, or within a single a head). Note the \emph{scoring} procedure is applied independently to each head and layer.

\subsection{Context-Calibrated Compression}
\label{method:cal_compress}

The above eviction procedure is parameterized in terms of $r \in (0, 1]$, the fraction of tokens in the KV cache to be retained. If one is given a token budget, the maximum $r$ is fully determined. If we have no token budget a priori, we would like to choose the {smallest} $r$ that leaves model quality unchanged. However, different contexts can tolerate wildly different compression ratios: a string of UUIDs is minimally compressible, while a context of few-shot examples may permit a higher compression. Table~\ref{tab:longbench} and Figure~\ref{fig:diff_perf} confirm this observation. To this end, we introduce \emph{context-calibrated compression}, a simple way of inferring the amount of compression (i.e number of tokens that can be evicted) that can be applied to a given text without performance on downstream tasks.

Let $t \in \mathcal V^{L}$ be a any text, and let $\LM(t_i ; {t}_{<i}, c)$ denote the language-model likelihood of the $i$-th token of $t$ conditioned on the preceding tokens in $t$ and any additional context $c$. To assess performance degradation, we would like a task-agnostic, scalar measure of how much compression degrades the model's output. A natural choice is the negative log-likelihood (NLL); note that a higher NLL indicates a more unlikely text. We can compute the token-averaged NLL for a given $t$ as follows
\begin{equation*}
    \operatorname{NLL}(t; c) = - \frac{1}{L}\sum_{i=1}^{L} \log(\LM(t_i ; {t}_{<i}, c))
\end{equation*}
Let $\widetilde c_r$ be the same context after applying a compression technique with retention $r$.  Let $q$ denote any question. Our goal is to predict, for any $(q, c,r)$ triple, the expected NLL ratio:
\[
  g(r,q,c) =
  \frac{\text{NLL}(t; q,c)}{\text{NLL}(t; q,\widetilde c_r)}
\]
for any text $t \in \mathcal V^{L}$. To make this possible in the query-agnostic setting we make the following crucial assumption: 
\[
  g(r,q,c) = \frac{\text{NLL}(t; q,c)}{\text{NLL}(t; q,\widetilde c_r)}\approx \frac{\text{NLL}(t;c)}{\text{NLL}(t;\widetilde c_r)} = f(r, c)
\]
for some function $f$, parameterized \emph{only} by $c, r$. I.e., we assume the gain in NLL of any text $t$ conditioned on $\tilde c_r$ can be explained entirely by the original context and $r$. Such an assumption is necessary in the query-agnostic regime. Empirically (see Fig.~\ref{fig:nll_curves}) we find  $g(r,c)$ decays smoothly and monotonically with $r$. As such, to approximate $g(r, q, c)\approx f(r, c)$, we fit the two-parameter curve
\[
  b=\alpha\cdot\text{NLL}(c)+\beta \qquad f_{\alpha,\beta}(r,c) =
  \frac{\exp\bigl(rb-b\bigr)-\exp(-b)}{1-\exp(-b)}  
\]
which satisfies $f_{\alpha,\beta}(1,c)=1$ (no compression) and
$f_{\alpha,\beta}(0,c)=0$ (empty cache). The free parameters $(\alpha,\beta)$ are obtained by least-squares regression
on training triples
\[
  \bigl(r,\,c,\,y\bigr)
  \qquad
  y=g(r, q, c)=\frac{\text{NLL}(t;q,c)}{\text{NLL}(t;q,\widetilde c_r)}
\]
where $q$ is the question being asked, $t$ is the ground truth answer, and $y$ is the ratio of the NLL of the true answer under the original and compressed contexts. At inference time, given a user-specified quality budget $\tau\in(0,1]$ and context $c$, (for example $\tau=0.95$ allows an NLL increase of at most $\frac{1}{0.95}\approx 5\%$), we choose $r^\star(c,\tau) = \arg\max_{r \in (0, 1)} f_{\alpha,\beta}(r,c) 
$ such that $f_{\alpha,\beta}(r,c) \ge\tau$, for which there is a closed form solution. \method then keeps the top $\lceil r^\star\!\cdot N\rceil$ tokens per head. In practice, we use  asymmetrical MSE to penalize under-estimation of $r^\star(c,\tau)$ (i.e weight such errors higher). In summary, we fit a curve to predict the level of degradation that will occur when retaining $r$ fraction of tokens under a particular compression method. At inference time we use the inverse of the curve to determine what retention rate the given context can support without excessively degrading performance. 

\section{Experiments}
  We demonstrate the efficacy of \method on two long-context models: \texttt{Llama 3.1-8B-Instruct} and \texttt{Qwen2.5-14B-Instruct-1M} (abbreviated as \texttt{Llama 3.1} and \texttt{Qwen 2.5} from here).  We first conduct benchmarking experiments to demonstrate the overhead cost of running the \method scoring mechanism. All benchmarks are conducted on an NVIDIA H100 80GB GPU, and we report median wall-clock time. We assess performance on long-context tasks and degradation across compression ratios via experiments on:

\noindent\textbf{RULER \citep{hsieh_ruler_2024}} is a synthetic benchmark that asks: \emph{given a stated context window, how much of it can the model actually use?} It consists of 13 tasks organized into four categories: (1) needle‑in‑a‑haystack (NIAH) retrieval with varying numbers of and types of ``needles'', (2) multi‑hop tracing that forces models to follow chains of references scattered through the text, (3) aggregation tasks that require fusing information from many locations, and (4) stress tests that add distractor tokens to short-form question tasks. We use the RULER-4k variant because (i) tested models demonstrate excellent performance with no compression, (ii) models have poor performance at higher compression rates, and (iii) we are chiefly concerned with performance as a function of \emph{compression rate}, instead of number of tokens retained. 

\noindent\textbf{Longbench \citep{bai_longbench_2024}} is a benchmark designed to evaluate long‐context understanding in LLMs. It consists of 21 datasets, covering the following task categories: single‐document QA, multi‐document QA, summarization, few‐shot learning, synthetic tasks, and code completion. LongBench requires comprehension of extensive passages, and probes models’ ability to maintain coherence, retrieve relevant information, and reason over non-synthetic long inputs. We evaluate on the English subset. 

Additionally, we assess the performance of context-calibrated compression on Longbench under two settings: Zero-Shot, and Finetuned. In the Zero Shot setting, for each document in the Longbench corpus, we predict the maximum compression rate the (eviction method, document) pair supports (see Sec ~\ref{sec:cal_compr}). In the Finetuned setting, we finetune the base model on documents in the Longbench corpus (not including queries or answers), and apply the aforementioned procedure to the finetuned model on the same Longbench corpus. Such an experiment is useful for assessing the compressive capability of the model when targeted for a particular corpus known a priori. The finetuning is performed with rank 128 LoRA adaptors on Q,K,V matrices for 3 epochs \citep{hu_lora_2022}.
\begin{wrapfigure}{r}{0.49\textwidth}
\centering
\begin{tikzpicture}[scale=0.70]
\begin{axis}[
    xlabel={Context Length},
    ylabel={Execution Time (ms)},
    xmin=4096, xmax=1048576,
    ymax=70,
    xmode=log,
    ymode=log,
    legend pos=south east,
    ymajorgrids=true,
    grid style=dashed,
    every axis plot/.append style={thick},
]

\addplot[
    blue!50,thick,mark=square,
    error bars/.cd,
    ]
    coordinates {
(256, 1.026687)(512, 1.032664)(1024, 1.046611)(2048, 1.081453)(4096, 1.145309)(8192, 1.274366)(16384, 1.518961)(32768, 2.001249)(65536, 2.975803)(131072, 5.902795)(262144, 7.768950)(524288, 19.506534)(1048576, 31.583626)
    };
        \addlegendentry{{\footnotesize {\method}}}   
            \addplot[
                thick,
    color=Bittersweet,
    mark=o,
    error bars/.cd,
    ]
    coordinates {
    (256, 1.345554)(512, 1.312034)(1024, 1.287826)(2048, 1.276906)(4096, 1.274009)(8192, 1.276523)(16384, 1.301756)(32768, 1.428951)(65536, 1.588616)(131072, 1.924723)(262144, 2.654723)(524288, 4.030933)(1048576, 6.812491)
    };
    \addlegendentry{{\footnotesize \method ($\vec o$)}}   
\addplot[
    thick,
    color=olive,
    mark=triangle,
    error bars/.cd,
    ]
    coordinates {
(256, 0.045867)(512, 0.053312)(1024, 0.069909)(2048, 0.103371)(4096, 0.196864)(8192, 0.393195)(16384, 0.786357)(32768, 1.558507)(65536, 3.116992)(131072, 6.309584)(262144, 9.866783)(524288, 19.913248)(1048576, 40.339840)
    };
        \addlegendentry{{\footnotesize {SnapKV}}}   

\addplot[
    thick,
    color=Gray,
    mark=pentagon,
    error bars/.cd,
    ]
    coordinates {
    (256, 0.030979)(512, 0.065325)(1024, 0.201217)(2048, 0.655040)(4096, 2.413425)(8192, 9.594114)(16384, 36.268608)(32768, 73.270481)
    };
    \addlegendentry{{\footnotesize $\textsf{H}_2\textsf{O}$}}

\addplot[
    thick,
    color=black,
    mark=diamond,
    error bars/.cd,
    ]
    coordinates {
(256, 0.023279)(512, 0.025610)(1024, 0.059699)(2048, 0.161176)(4096, 0.490879)(8192, 1.690028)(16384, 6.244441)(32768, 23.929577)(65536, 95.862877)
    };
        \addlegendentry{{\footnotesize {FlashAttn} 2 }}   
\end{axis}
\end{tikzpicture}

\caption{
 Median wall-clock overhead of the selection mechanism for one layer of \texttt{Llama 3.1} (batch size 1). }\label{fig:benchmark}
\vspace{1.3em}
\begin{tikzpicture}[scale=0.75]
  \begin{axis}[
      ybar,
      bar width=12pt,
      width=9cm,
      height=6cm,                  
      ymin=0,
      ymax=1.1,
      ylabel={Performance (\% of Full KV)},
      symbolic x coords={TOVA,KeyDiff,SnapKV,\colorbox{LightCyan}{\method}},
      xtick=data,
      ytick={0,0.2,0.4,0.6,0.8, 1.0},
      yticklabels={0, 20\%, 40\%, 60\%, 80\%, 100\%},
      ymajorgrids=true,
      x tick label style={yshift=-2pt,rotate=25,anchor=east},
      legend pos=north west,
      legend cell align={left},
      axis y line=left,
      axis x line=bottom,
      axis on top=true,
     enlarge x limits=0.35,
    ]
    \addplot+[bar shift=-7pt, draw=black, pattern=north west lines,pattern color=PineGreen]
      coordinates {
        (TOVA,0.0265)
        (KeyDiff,0.8817)
        (SnapKV,0.478)
        (\colorbox{LightCyan}{\method},0.988)
      };
    \addplot+[bar shift=7pt, draw=black, pattern=crosshatch dots, pattern color=blue]
      coordinates {
        (TOVA,0.8897742364)
        (KeyDiff,0.89600000)
        (SnapKV,0.8645418327)
        (\colorbox{LightCyan}{\method},0.956)
      };
    \legend{\texttt{NIAH}, \texttt{QA}}
  \end{axis}
\end{tikzpicture}
    \vspace{-1em}
    \caption{Performance of eviction methods on RULER NIAH UUID and QA subtasks at  50\% KV retention cache on \texttt{Llama 3.1}.}
    \vspace{-4.5em}
    \label{fig:diff_perf}
\end{wrapfigure}

We use the following parameters for all experiments: $\lambda=0.3$, sketching dimension $k=48$, chunk size $C=256$, $\tau=0.95$ for \texttt{Llama 3.1} and $\tau=0.90$ for \texttt{Qwen 2.5}.  We demonstrate the stability of \method across parameter choices in \S~\ref{sec:ablation}. While the choice of $k=48$ violates the bounds required for Theorem~\ref{thm:approx_lev_scores}, we empirically find the approximation is sufficient to achieve good performance. Such a phenomenon is not a new observation, and much prior work has shown that the bounds tend to be loose on real-world datasets \citep{broadbent_subset_2010,paschou_pca-correlated_2007, jolliffe_discarding_1972}. All experiments are conducted in a query-agnostic manner, and with head-adaptive compression \citep{feng_ada-kv_2025} (except \ref{fig:nll_curves}, which is computed without head-adaptive compression to better qualitatively demonstrate task-variability). We compare \method to TOVA \citep{oren_transformers_2024}, KeyDiff \citep{park_keydiff_2025}, SnapKV \citep{li_snapkv_2024}, DuoAttention \citep{xiao_duoattention_2024}, and PyramidKV \citep{cai_pyramidkv_2024} as baselines. We exclude $\textsf{H}_2\textsf{O}$ given its computational expense (see Figure~\ref{fig:benchmark}). We use greedy sampling for all results. 

\section{Results}
\subsection{Selection Mechanism Speed}

We benchmark the speed of different eviction mechanisms across various context lengths. Figure~\ref{fig:benchmark} shows wall-clock execution time of the eviction mechanism for a single layer for \texttt{Llama 3.1} (with batch size 1). We plot the execution time of Flash Attention 2 to provide context as a lower bound on the amount of time spent in a layer. We see that \method is about as fast as SnapKV for context lengths above $16$k tokens; the overhead at short contexts is due to the time spent computing an SVD, but this time is constant for all context lengths.

We additionally show the cost of computing only outlier scores 
(\method ($\vec o$)) to demonstrate the minimal overhead of approximate leverage score computation; the increase in execution time for \method (no $\vec a$) is almost entirely due to matrix multiplication (line 5 in Algorithm~\ref{alg:leverage-sampling}).
As such, \method can be used as a drop-in replacement over other eviction mechanisms in the query-agnostic regime, with very little performance overhead for long contexts.

\subsection{RULER}
\begin{figure}[h]
\vspace{-.5em}
\centering
\begin{tikzpicture}[scale=0.75]
  \begin{groupplot}[
      group style={group size=2 by 1, horizontal sep=1.5cm},
      width=7cm,    
      height=6cm,   
      xlabel={KV Retention (\%)},
      xmin=0, xmax=100,
      x dir=reverse,
      ymajorgrids=true,
      grid style=dashed,
        legend columns=7,                     
              legend style={
                at={(1.075,-0.25)}, anchor=north, 
                /tikz/every even column/.append style={column sep=0.7em},
              },
      every axis plot/.append style={thick},
    ]
\
    \nextgroupplot[
      title={\texttt{Llama 3.1}},
      ylabel={Mean RULER Score},
      xlabel={KV Retention ($\%$)},
    ]
      \addplot[color=black, domain=0:100, samples=2, dashed]{95.57876587};
      \addlegendentry{Base}

      \addplot[blue!50, thick, mark=square] coordinates {
        (100, 95.316154) (75, 95.626154) (50, 94.7)
        (25, 83.075385) (10, 64.793077)};
      \addlegendentry{Compactor}

      \addplot[thick, color=olive, mark=triangle] coordinates {
        (100, 95.57876587) (75, 88.130769) (50, 77.600769)
        (25, 63) (10, 43)};
      \addlegendentry{SnapKV}

      \addplot[thick, color=Gray, mark=pentagon] coordinates {
        (100, 95.57876587) (75, 87.364615) (50, 76.214615)
        (25, 63.370000) (10, 37.610000)};
      \addlegendentry{TOVA}

      \addplot[thick, color=pink, mark=diamond] coordinates {
        (100, 95.57876587) (75, 88.130769) (50, 77.643846)
        (25, 48.679231) (10, 27.753077)};
      \addlegendentry{PyramidKV}

      \addplot[color=Bittersweet, thick, mark=o] coordinates {
        (100, 95.57876587) (75, 63.650769) (50, 7.386154)
        (25, 0.472308) (10, 0.149231)};
      \addlegendentry{Random}

      \addplot[color=orange, thick, mark=+] coordinates {
        (100, 95.57876587) (75, 94.3) (50, 90.83782)
        (25, 73.197692) (10, 24.514615)};
      \addlegendentry{DuoAttention}

    \nextgroupplot[
      title={\texttt{Qwen 2.5}},
      xlabel={KV Retention ($\%$)},
    ]
      \addplot[color=black, domain=0:100, samples=2, dashed]{95.67823029};

      \addplot[blue!50, thick, mark=square] coordinates {
        (100, 95.67823029) (75, 96.150769) (50, 95.496923)
        (25, 83.474615) (10, 66.186154)};

      \addplot[thick, color=olive, mark=triangle] coordinates {
        (100, 95.67823029) (75, 92.370769) (50, 75.252308)
        (25, 46.027692) (10, 28.919231)};

      \addplot[thick, color=Gray, mark=pentagon] coordinates {
        (100, 95.67823029) (75, 83.142308) (50, 79.206923)
        (25, 65.951538) (10, 25.360769)};

      \addplot[thick, color=pink, mark=diamond] coordinates {
        (100, 95.67823029) (75, 92.393077) (50, 75.286154)
        (25, 46.027692) (10, 28.880769)};

      \addplot[color=Bittersweet, thick, mark=o] coordinates {
        (100, 95.67823029) (75, 83.142308) (50, 39.5402832)
        (25, 23.69850159) (10, 16.34572029)};

  \end{groupplot}
\end{tikzpicture}
\caption{Mean RULER score (all tasks) on \texttt{Llama 3.1} and \texttt{Qwen 2.5} across KV retention rates.}
\label{fig:ruler_graph}
\vspace{-1em}
\end{figure}
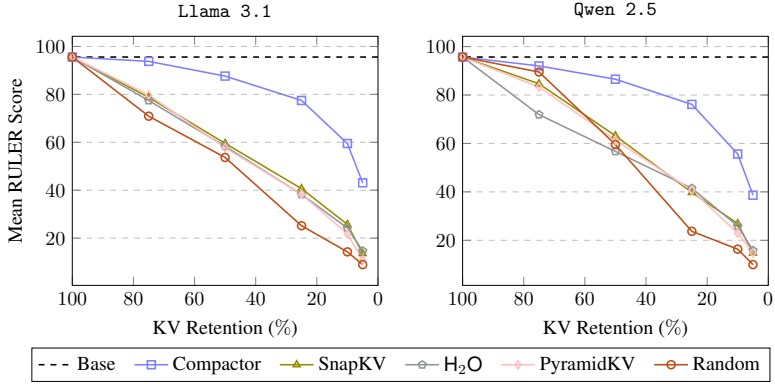
Figure~\ref{fig:ruler_graph} plots mean RULER score (across all tasks) versus KV retention for \method and several baselines (SnapKV,  TOVA, KeyDiff, PyramidKV, DuoAttention, random eviction) for both \texttt{Llama 3.1} and \texttt{Qwen 2.5} (no checkpoint for DuoAttention is available for \texttt{Qwen 2.5}).  $100\%$ KV retention denotes the score of the model with no compression and is shown as a dashed line.  We observe remarkably similar performance characteristics across both models. As the cache is pruned, \method degrades most gracefully: at  $50\%$ retention it recovers  $99\%$ of baseline, at  $25\% $ it maintains  $87.1\% $, and even at only  $10\% $ retention it still achieves  $68.0\%$. Performance when using SnapKV, TOVA, PyramidKV, and DuoAttention degrades poorly with KV retention setting, confirming the inadequacy of these methods in the query-agnostic regime. The performance of all methods is poor at severe compression ratios, demonstrating the inherent difficulty of the task. Performance of eviction methods on NIAH and QA tasks is shown in Figure~\ref{fig:diff_perf}, and demonstrates the variable performance across different tasks, which further motivates the need for context-calibrated compression.

\begin{table}[h]
  \centering
  \vspace{-.2em}
  \caption{Longbench scores by meta-task. 
  Cells in blue denote our contribution.}
  \vspace{.2em}
  \scalebox{0.75}{
  \begin{tabular}{llcccccccccc}
    \toprule
    \multirow{4}{*}{Task} &\multirow{4}{*}{Method}& \multicolumn{5}{c}{\texttt{Llama 3.1}} & \multicolumn{5}{c}{\texttt{Qwen 2.5}}\\
    \cmidrule(r){3-7}\cmidrule(r){8-12}

    & & 100\% & 50\% & 25\% & 10\% & 5\% & 100\% & 50\% & 25\% & 10\% & 5\% \\
    \midrule
    \multirow{4}{*}{Single-Doc QA} & \method & \multirow{4}{*}{\textit{44.6}} &{\cellcolor{LightCyan}}\textbf{44.5} & {\cellcolor{LightCyan}}\textbf{41.9} & {\cellcolor{LightCyan}}\textbf{35.8} &{\cellcolor{LightCyan}} \textbf{30.0} & \multirow{4}{*}{\textit{42.0}} & {\cellcolor{LightCyan}}\textbf{41.6} & {\cellcolor{LightCyan}}\textbf{38.3} & {\cellcolor{LightCyan}}\textbf{33.3} & {\cellcolor{LightCyan}}\textbf{25.3}\\
    & KeyDiff &  & 44.3 & 39.5 & 30.2 & 22.2 & &41.3 & 36.2 & 25.7 & 19.3\\
    & SnapKV & & 40.6 & 33.4 & 25.3 & 18.9 &&37.4 & 29.8 & 22.5 & 17.7\\
    & TOVA & & 41.8 & 32.7 & 22.1 & 15.8 && 38.4 & 29.9 & 19.2 & 15.4\\
    \midrule

    \multirow{4}{*}{Multi-Doc QA} & \method & \multirow{4}{*}{\textit{39.3}} & {\cellcolor{LightCyan}}\textbf{38.8} & {\cellcolor{LightCyan}}\textbf{36.4} & {\cellcolor{LightCyan}}\textbf{32.8} & {\cellcolor{LightCyan}}\textbf{27.9} & \multirow{4}{*}{\textit{46.1}} & {\cellcolor{LightCyan}}\textbf{45.1} & {\cellcolor{LightCyan}}\textbf{45.4} & {\cellcolor{LightCyan}}\textbf{39.4} & {\cellcolor{LightCyan}}\textbf{33.5}\\
    & KeyDiff && 37.8 & 33.7 & 25.2 & 20.1 & & 43.1 & 38.0 & 28.4 & 22.5\\
    & SnapKV && 36.7 & 32.53 & 24.8 & 20.9 & & 42.8 & 38.5 & 29.4 & 24.5\\
    & TOVA && 36.7 & 34.2 & 25.6 & 20.6 && 44.5 & 41.4 & 31.7 & 24.5\\
    \midrule

    \multirow{4}{*}{Summarization} & \method & \multirow{4}{*}{\textit{28.7}} & {\cellcolor{LightCyan}}\textbf{28.2} & {\cellcolor{LightCyan}}\textbf{26.9} & {\cellcolor{LightCyan}}\textbf{25.0} & {\cellcolor{LightCyan}}\textbf{23.3} & \multirow{4}{*}{\textit{27.8}} & {\cellcolor{LightCyan}}\textbf{27.5} & {\cellcolor{LightCyan}}\textbf{26.9} & {\cellcolor{LightCyan}}\textbf{25.7} & {\cellcolor{LightCyan}}\textbf{23.7}\\
    & KeyDiff & &\textbf{28.2} & 26.7 & 23.4 & 21.3 & & 27.2 & 25.3 & 22.1 & 19.9\\
    & SnapKV & & 27.1 & 24.9 & 22.5 & 20.2 & & 26.5 & 24.6 & 22.3 & 20.0\\
    & TOVA && 27.0 & 25.0 & 22.0 & 20.2 & & 26.6 &24.7 & 21.8 & 19.6\\
            \midrule

    \multirow{4}{*}{Few Shot} & \method & \multirow{4}{*}{\textit{48.3}} & {\cellcolor{LightCyan}}\textbf{58.9} & {\cellcolor{LightCyan}}{54.8} & {\cellcolor{LightCyan}}\textbf{50.0} & {\cellcolor{LightCyan}}{42.2} & \multirow{4}{*}{\textit{65.6}} & {\cellcolor{LightCyan}}\textbf{64.4} & {\cellcolor{LightCyan}}\textbf{63.1} & {\cellcolor{LightCyan}}{51.3}& {\cellcolor{LightCyan}}{45.4}\\
    & KeyDiff& & 56.3 & \textbf{54.9} & 49.9 & 44.1 & &\textbf{64.4} & 61.5 & 53.8 & 46.5\\
    & SnapKV& & 50.8 & 53.0 & 48.4 & \textbf{45.1} & &\textbf{64.4} & 62.9 & \textbf{56.4} & \textbf{50.8}\\
    & TOVA && 44.1 & 42.2 & 40.0 & 40.3 & & 64.2 & 58.9 & 51.0 & 46.7 \\
            \midrule

    \multirow{4}{*}{Code} & \method & \multirow{4}{*}{\textit{43.1}} & {\cellcolor{LightCyan}}\textbf{45.1} & {\cellcolor{LightCyan}}\textbf{44.4} & {\cellcolor{LightCyan}}\textbf{44.0} & {\cellcolor{LightCyan}}{42.9} & \multirow{4}{*}{\textit{62.4}} & {\cellcolor{LightCyan}}\textbf{62.2} & {\cellcolor{LightCyan}}{61.2} & {\cellcolor{LightCyan}}{55.6}& {\cellcolor{LightCyan}}{53.2}\\
    & KeyDiff& & 44.0 & 37.4 & 32.0 & 29.7 & & 58.73 & 46.3 &37.1 & 36.2\\
    & SnapKV& & 41.4 & 41.9 & 43.8 & \textbf{46.2}& & {62.1} & {61.3} & \textbf{60.0} & \textbf{59.1}\\
    & TOVA && 43.4 & 43.4 & 44.0 & 45.8 && \textbf{62.2} & \textbf{61.9} & 59.4 & 58.3\\
            \midrule
    \multirow{4}{*}{Total} & \method & \multirow{4}{*}{\textit{42.4}} & {\cellcolor{LightCyan}}\textbf{44.9} & {\cellcolor{LightCyan}}\textbf{42.5} & {\cellcolor{LightCyan}}\textbf{38.8} & {\cellcolor{LightCyan}}\textbf{33.9} & \multirow{4}{*}{\textit{50.1}} & {\cellcolor{LightCyan}}\textbf{49.4} & {\cellcolor{LightCyan}}\textbf{48.1} & {\cellcolor{LightCyan}}\textbf{41.8} & {\cellcolor{LightCyan}}\textbf{36.6} \\
    & KeyDiff& & 43.9 & 40.0& 33.1 & 27.7 & & 48.3 & 42.9 & 34.2 & 29.4\\
    & SnapKV& & 41.2 & 38.5 & 33.3 & 29.7 & & 47.9 & 44.1 & 38.0 & 33.6\\
    & TOVA && 40.2 & 36.9 & 30.9 & 28.1  &&  48.4 & 44.3 & 36.6 & 31.9\\
    \bottomrule
  \end{tabular}}\label{tab:longbench}
  \vspace{-1.5em}
\end{table}
\subsection{Longbench}

We evaluate \method on Longbench to assess performance on real-world long context tasks. Table~\ref{tab:longbench} reports per-task and average scores on LongBench for \method, TOVA, KeyDiff, and SnapKV at  $50\% $,  $25\% $,  $10\% $, and  $5\%$ retention. PyramidKV was excluded due to how similarly it performs to SnapKV. At  $50\%$ retention, \method matches or slightly exceeds the full-cache performance, with particular gains in few-shot and coding tasks with less contextual diversity.  Even at  $25\% $, \method retains around $95\%$ of the full-cache score.  Under more aggressive pruning ($10\% $ and  $5\%$), \method’s advantage grows, outperforming competing methods by over 15\%. Furthermore, \method performs well for all task categories, whereas some methods greatly underperform (e.g KeyDiff on Code tasks, or TOVA on Single-Doc QA tasks). Such task-agnostic performance is essential in practical deployments of KV cache compression. These results confirm that \method reliably preserves tokens that drive task performance.

\subsection{Context-Calibrated Compression}\label{sec:cal_compr}

\begin{figure}[b]
\vspace{-1em}
    \centering
\begin{tikzpicture}[scale=0.8]
  \begin{groupplot}[
      group style={
        group size=2 by 1,
        vertical sep=1.5cm,
        horizontal sep=2cm
      },
      width=12cm, height=5cm,
      xmin=0, xmax=100,
      x dir=reverse,                  
      ymin=0, ymax=1.1,
      ytick={0,0.5,1},
      legend pos=south west,
      tick align=outside,
      width=0.55\linewidth
    ]
    \nextgroupplot[
      title={\texttt{Llama 3.1}},
      ylabel={NLL Ratio},
      xlabel={KV Retention ($\%$)},
    ]
            \coordinate (curvept_compactor) at (65,0.65);
        \coordinate (curvept_snapkv) at (55,0.43);
        \draw[->, thick]
      (80,.65) node[anchor=east]{\tiny UUID task}
      -- (curvept_compactor);
        \draw[->, thick]
      (80,.65) node[anchor=east]{}
      -- (curvept_snapkv);
\addplot[solid,blue!50] coordinates {
  (100.0,1.0) (95.0,1.0) (90.0,1.0) (85.0,1.0) (80.0,1.0)
  (75.0,1.0) (70.0,1.0) (65.0,1.0) (60.0,1.0) (55.0,1.0)
  (50.0,1.0) (45.0,1.0) (40.0,1.0) (35.0,1.0) (30.0,1.0)
  (25.0,1.0) (20.0,1.0) (15.0,1.0) (10.0,1.0) (5.0,0.97) (0,0)
};
\addlegendentry{Compactor}
\addplot[solid,blue!50,forget plot] coordinates {
  (100.0,1.0) (95.0,1.0) (90.0,1.0) (85.0,1.0) (80.0,1.0)
  (75.0,1.0) (70.0,1.0) (65.0,1.0) (60.0,1.0) (55.0,1.0)
  (50.0,1.0) (45.0,1.0) (40.0,1.0) (35.0,1.0) (30.0,1.0)
  (25.0,1.0) (20.0,1.0) (15.0,1.0) (10.0,1.0) (5.0,0.97) (0,0)
};
\addplot[solid,blue!50,forget plot] coordinates {
  (100.0,1.0) (95.0,1.0) (90.0,1.0) (85.0,1.0) (80.0,1.0)
  (75.0,1.0) (70.0,1.0) (65.0,1.0) (60.0,1.0) (55.0,1.0)
  (50.0,1.0) (45.0,1.0) (40.0,1.0) (35.0,1.0) (30.0,1.0)
  (25.0,1.0) (20.0,0.99) (15.0,0.97) (10.0,0.91) (5.0,0.70) (0,0)
};
\addplot[solid,blue!50,forget plot] coordinates {
  (100.0,1.0) (95.0,1.0) (90.0,1.0) (85.0,1.0) (80.0,1.0)
  (75.0,1.0) (70.0,1.0) (65.0,1.0) (60.0,1.0) (55.0,1.0)
  (50.0,1.0) (45.0,1.0) (40.0,1.0) (35.0,1.0) (30.0,1.0)
  (25.0,1.0) (20.0,1.0) (15.0,1.0) (10.0,0.99) (5.0,0.91) (0,0)
};
\addplot[solid,blue!50,forget plot] coordinates {
  (100.0,1.0) (95.0,1.0) (90.0,1.0) (85.0,1.0) (80.0,1.0)
  (75.0,1.0) (70.0,1.0) (65.0,1.0) (60.0,1.0) (55.0,1.0)
  (50.0,1.0) (45.0,1.0) (40.0,1.0) (35.0,1.0) (30.0,1.0)
  (25.0,1.0) (20.0,1.0) (15.0,0.98) (10.0,0.93) (5.0,0.74) (0,0)
};
\addplot[solid,blue!50,forget plot] coordinates {
  (100.0,1.0) (95.0,1.0) (90.0,1.0) (85.0,1.0) (80.0,1.0)
  (75.0,1.0) (70.0,1.0) (65.0,1.0) (60.0,1.0) (55.0,1.0)
  (50.0,1.0) (45.0,1.0) (40.0,1.0) (35.0,1.0) (30.0,0.99)
  (25.0,0.98) (20.0,0.95) (15.0,0.90) (10.0,0.78) (5.0,0.53) (0,0)
};
\addplot[solid,blue!50,forget plot] coordinates {
  (100.0,1.0) (95.0,1.0) (90.0,1.0) (85.0,1.0) (80.0,1.0)
  (75.0,1.0) (70.0,1.0) (65.0,1.0) (60.0,1.0) (55.0,1.0)
  (50.0,1.0) (45.0,1.0) (40.0,1.0) (35.0,1.0) (30.0,0.99)
  (25.0,0.98) (20.0,0.95) (15.0,0.90) (10.0,0.78) (5.0,0.53) (0,0)
};
\addplot[solid,blue!50,forget plot] coordinates {
  (100.0,1.0) (95.0,1.0) (90.0,1.0) (85.0,1.0) (80.0,1.0)
  (75.0,1.0) (70.0,1.0) (65.0,1.0) (60.0,1.0) (55.0,1.0)
  (50.0,1.0) (45.0,1.0) (40.0,1.0) (35.0,1.0) (30.0,0.99)
  (25.0,0.98) (20.0,0.95) (15.0,0.90) (10.0,0.79) (5.0,0.54) (0,0)
};
\addplot[solid,blue!50,forget plot] coordinates {
  (100.0,1.0) (95.0,1.0) (90.0,1.0) (85.0,1.0) (80.0,1.0)
  (75.0,1.0) (70.0,1.0) (65.0,1.0) (60.0,1.0) (55.0,1.0)
  (50.0,1.0) (45.0,1.0) (40.0,1.0) (35.0,1.0) (30.0,1.0)
  (25.0,1.0) (20.0,1.0) (15.0,1.0) (10.0,0.97) (5.0,0.84) (0,0)
};
\addplot[solid,blue!50,forget plot] coordinates {
  (100.0,1.0) (95.0,1.0) (90.0,1.0) (85.0,1.0) (80.0,1.0)
  (75.0,1.0) (70.0,1.0) (65.0,1.0) (60.0,1.0) (55.0,1.0)
  (50.0,1.0) (45.0,1.0) (40.0,1.0) (35.0,1.0) (30.0,1.0)
  (25.0,1.0) (20.0,1.0) (15.0,0.99) (10.0,0.95) (5.0,0.78) (0,0)
};
\addplot[solid,blue!50,forget plot] coordinates {
  (100.0,1.0) (95.0,1.0) (90.0,1.0) (85.0,1.0) (80.0,1.0)
  (75.0,1.0) (70.0,1.0) (65.0,1.0) (60.0,1.0) (55.0,1.0)
  (50.0,1.0) (45.0,1.0) (40.0,1.0) (35.0,1.0) (30.0,1.0)
  (25.0,1.0) (20.0,1.0) (15.0,1.0) (10.0,1.0) (5.0,0.97) (0,0)
};
\addplot[solid,blue!50,forget plot] coordinates {
  (100.0,1.0) (95.0,1.0) (90.0,1.0) (85.0,1.0) (80.0,1.0)
  (75.0,1.0) (70.0,1.0) (65.0,1.0) (60.0,1.0) (55.0,1.0)
  (50.0,1.0) (45.0,1.0) (40.0,1.0) (35.0,1.0) (30.0,0.99)
  (25.0,0.98) (20.0,0.95) (15.0,0.90) (10.0,0.79) (5.0,0.54) (0,0)
};
\addplot[solid,blue!50,forget plot] coordinates {
  (100.0,1.0) (95.0,0.95) (90.0,0.91) (85.0,0.86) (80.0,0.82)
  (75.0,0.77) (70.0,0.72) (65.0,0.67) (60.0,0.63) (55.0,0.58)
  (50.0,0.53) (45.0,0.48) (40.0,0.43) (35.0,0.37) (30.0,0.32)
  (25.0,0.27) (20.0,0.22) (15.0,0.16) (10.0,0.11) (5.0,0.06) (0,0)
};
\addplot[solid,blue!50,forget plot] coordinates {
  (100.0,1.0) (95.0,1.0) (90.0,1.0) (85.0,1.0) (80.0,1.0)
  (75.0,1.0) (70.0,1.0) (65.0,1.0) (60.0,1.0) (55.0,1.0)
  (50.0,1.0) (45.0,1.0) (40.0,1.0) (35.0,0.99) (30.0,0.99)
  (25.0,0.98) (20.0,0.95) (15.0,0.89) (10.0,0.78) (5.0,0.53) (0,0)
};
    \addplot[dashed,red] coordinates {
  (100.0,1.0) (89.0,1.0) (79.0,1.0) (68.0,1.0)
  (58.0,1.0) (47.0,1.0) (37.0,1.0) (26.0,1.0)
  (16.0,0.96) (5.0,0.66) (0,0)
};
    \addlegendentry{SnapKV}
\addplot[dashed,red,forget plot] coordinates {
  (100.0,1.0) (89.0,1.0) (79.0,1.0) (68.0,0.99)
  (58.0,0.98) (47.0,0.95) (37.0,0.90) (26.0,0.81)
  (16.0,0.63) (5.0,0.27) (0,0)
};
\addplot[dashed,red,forget plot] coordinates {
  (100.0,1.0) (89.0,1.0) (79.0,1.0) (68.0,1.0)
  (58.0,1.0) (47.0,1.0) (37.0,0.99) (26.0,0.97)
  (16.0,0.88) (5.0,0.49) (0,0)
};
\addplot[dashed,red,forget plot] coordinates {
  (100.0,1.0) (89.0,1.0) (79.0,1.0) (68.0,0.99)
  (58.0,0.98) (47.0,0.97) (37.0,0.93) (26.0,0.85)
  (16.0,0.67) (5.0,0.30) (0,0)
};
\addplot[dashed,red,forget plot] coordinates {
  (100.0,1.0) (89.0,0.99) (79.0,0.97) (68.0,0.94)
  (58.0,0.90) (47.0,0.84) (37.0,0.76) (26.0,0.63)
  (16.0,0.44) (5.0,0.17) (0,0)
};
\addplot[dashed,red,forget plot] coordinates {
  (100.0,1.0) (89.0,0.99) (79.0,0.97) (68.0,0.94)
  (58.0,0.90) (47.0,0.84) (37.0,0.75) (26.0,0.63)
  (16.0,0.44) (5.0,0.17) (0,0)
};
\addplot[dashed,red,forget plot] coordinates {
  (100.0,1.0) (89.0,0.99) (79.0,0.97) (68.0,0.94)
  (58.0,0.90) (47.0,0.85) (37.0,0.76) (26.0,0.63)
  (16.0,0.45) (5.0,0.17) (0,0)
};
\addplot[dashed,red,forget plot] coordinates {
  (100.0,1.0) (89.0,1.0) (79.0,1.0) (68.0,1.0)
  (58.0,1.0) (47.0,0.99) (37.0,0.97) (26.0,0.93)
  (16.0,0.79) (5.0,0.39) (0,0)
};
\addplot[dashed,red,forget plot] coordinates {
  (100.0,1.0) (89.0,1.0) (79.0,1.0) (68.0,1.0)
  (58.0,0.99) (47.0,0.98) (37.0,0.95) (26.0,0.88)
  (16.0,0.72) (5.0,0.34) (0,0)
};
\addplot[dashed,red,forget plot] coordinates {
  (100.0,1.0) (89.0,1.0) (79.0,1.0) (68.0,1.0)
  (58.0,1.0) (47.0,1.0) (37.0,1.0) (26.0,1.0)
  (16.0,0.96) (5.0,0.64) (0,0)
};
\addplot[dashed,red,forget plot] coordinates {
  (100.0,1.0) (89.0,0.99) (79.0,0.97) (68.0,0.94)
  (58.0,0.90) (47.0,0.85) (37.0,0.76) (26.0,0.64)
  (16.0,0.45) (5.0,0.17) (0,0)
};
\addplot[dashed,red,forget plot] coordinates {
  (100.0,1.0) (89.0,0.85) (79.0,0.71) (68.0,0.58)
  (58.0,0.46) (47.0,0.36) (37.0,0.27) (26.0,0.18)
  (16.0,0.10) (5.0,0.03) (0,0)
};
\addplot[dashed,red,forget plot] coordinates {
  (100.0,1.0) (89.0,0.99) (79.0,0.97) (68.0,0.94)
  (58.0,0.90) (47.0,0.84) (37.0,0.75) (26.0,0.62)
  (16.0,0.44) (5.0,0.17) (0,0)
};
    \nextgroupplot[
      title={\texttt{Qwen 2.5}},
      xlabel={KV Retention ($\%$)},
    ]
                \coordinate (curvept_compactor) at (59,0.56);
        \draw[->, thick]
      (80,.25) node[anchor=east]{\tiny UUID task}
      -- (curvept_compactor);
    \addplot[dashed,red] coordinates {
(100.0, 1.0) (89.0, 1.0) (79.0, 1.0) (68.0, 1.0) (58.00000000000001, 0.99) (47.0, 0.99) (37.0, 0.96) (26.0, 0.9) (16.000000000000004, 0.75) (5.000000000000004, 0.36) (0, 0)
    };
\addplot[dashed,red,forget plot] coordinates {
  (100.0, 1.0) (89.0, 0.97) (79.0, 0.94) (68.0, 0.90)
  (58.0, 0.84) (47.0, 0.76) (37.0, 0.66) (26.0, 0.53)
  (16.0, 0.36) (5.0, 0.13) (0, 0)
};
\addplot[dashed,red,forget plot] coordinates {
  (100.0, 1.0) (89.0, 0.99) (79.0, 0.99) (68.0, 0.97)
  (58.0, 0.95) (47.0, 0.91) (37.0, 0.84) (26.0, 0.72)
  (16.0, 0.53) (5.0, 0.22) (0, 0)
};
\addplot[dashed,red,forget plot] coordinates {
  (100.0, 1.0) (89.0, 0.99) (79.0, 0.97) (68.0, 0.94)
  (58.0, 0.90) (47.0, 0.84) (37.0, 0.76) (26.0, 0.63)
  (16.0, 0.45) (5.0, 0.17) (0, 0)
};
\addplot[dashed,red,forget plot] coordinates {
  (100.0, 1.0) (89.0, 0.86) (79.0, 0.74) (68.0, 0.62)
  (58.0, 0.50) (47.0, 0.40) (37.0, 0.30) (26.0, 0.21)
  (16.0, 0.12) (5.0, 0.04) (0, 0)
};
\addplot[dashed,red,forget plot] coordinates {
  (100.0, 1.0) (89.0, 0.85) (79.0, 0.72) (68.0, 0.60)
  (58.0, 0.48) (47.0, 0.38) (37.0, 0.28) (26.0, 0.19)
  (16.0, 0.11) (5.0, 0.03) (0, 0)
};
\addplot[dashed,red,forget plot] coordinates {
  (100.0, 1.0) (89.0, 0.86) (79.0, 0.73) (68.0, 0.61)
  (58.0, 0.50) (47.0, 0.40) (37.0, 0.30) (26.0, 0.20)
  (16.0, 0.12) (5.0, 0.04) (0, 0)
};
\addplot[dashed,red,forget plot] coordinates {
  (100.0, 1.0) (89.0, 0.99) (79.0, 0.98) (68.0, 0.96)
  (58.0, 0.93) (47.0, 0.88) (37.0, 0.80) (26.0, 0.68)
  (16.0, 0.49) (5.0, 0.20) (0, 0)
};
\addplot[dashed,red,forget plot] coordinates {
  (100.0, 1.0) (89.0, 0.98) (79.0, 0.95) (68.0, 0.90)
  (58.0, 0.85) (47.0, 0.77) (37.0, 0.68) (26.0, 0.54)
  (16.0, 0.37) (5.0, 0.14) (0, 0)
};
\addplot[dashed,red,forget plot] coordinates {
  (100.0, 1.0) (89.0, 1.00) (79.0, 1.00) (68.0, 1.00)
  (58.0, 0.99) (47.0, 0.98) (37.0, 0.96) (26.0, 0.89)
  (16.0, 0.74) (5.0, 0.35) (0, 0)
};
\addplot[dashed,red,forget plot] coordinates {
  (100.0, 1.0) (89.0, 0.87) (79.0, 0.75) (68.0, 0.63)
  (58.0, 0.52) (47.0, 0.41) (37.0, 0.31) (26.0, 0.22)
  (16.0, 0.12) (5.0, 0.04) (0, 0)
};
\addplot[dashed,red,forget plot] coordinates {
  (100.0, 1.0) (89.0, 0.91) (79.0, 0.81) (68.0, 0.71)
  (58.0, 0.61) (47.0, 0.51) (37.0, 0.40) (26.0, 0.29)
  (16.0, 0.17) (5.0, 0.06) (0, 0)
};
\addplot[dashed,red,forget plot] coordinates {
  (100.0, 1.0) (89.0, 0.86) (79.0, 0.74) (68.0, 0.62)
  (58.0, 0.51) (47.0, 0.40) (37.0, 0.30) (26.0, 0.21)
  (16.0, 0.12) (5.0, 0.04) (0, 0)
};

    \addplot[ solid,blue!50] coordinates {
(100.0, 1.0) (89.0, 1.0) (79.0, 1.0) (68.0, 1.0) (58.00000000000001, 1.0) (47.0, 0.99) (37.0, 0.98) (26.0, 0.93) (16.000000000000004, 0.8) (5.000000000000004, 0.41) (0, 0)
    };
    \addplot[ solid,blue!50,forget plot] coordinates {
(100.0, 1.0) (89.0, 1.0) (79.0, 1.0) (68.0, 0.99) (58.00000000000001, 0.98) (47.0, 0.95) (37.0, 0.9) (26.0, 0.81) (16.000000000000004, 0.63) (5.000000000000004, 0.27) (0, 0)
    };

        \addplot[ solid,blue!50,forget plot] coordinates {
(100.0, 1.0) (89.0, 1.0) (79.0, 1.0) (68.0, 1.0) (58.00000000000001, 0.99) (47.0, 0.97) (37.0, 0.94) (26.0, 0.87) (16.000000000000004, 0.7) (5.000000000000004, 0.32) (0, 0)
    };
        \addplot[ solid,blue!50,forget plot] coordinates {
(100.0, 1.0) (89.0, 1.0) (79.0, 1.0) (68.0, 0.99) (58.00000000000001, 0.98) (47.0, 0.96) (37.0, 0.92) (26.0, 0.84) (16.000000000000004, 0.66) (5.000000000000004, 0.3) (0, 0)
    };    \addplot[ solid,blue!50,forget plot] coordinates {
(100.0, 1.0) (89.0, 0.99) (79.0, 0.98) (68.0, 0.96) (58.00000000000001, 0.93) (47.0, 0.88) (37.0, 0.8) (26.0, 0.68) (16.000000000000004, 0.49) (5.000000000000004, 0.2) (0, 0)
    };    \addplot[ solid,blue!50,forget plot] coordinates {
(100.0, 1.0) (89.0, 0.99) (79.0, 0.98) (68.0, 0.96) (58.00000000000001, 0.92) (47.0, 0.87) (37.0, 0.8) (26.0, 0.67) (16.000000000000004, 0.49) (5.000000000000004, 0.19) (0, 0)
    };    \addplot[ solid,blue!50,forget plot] coordinates {
(100.0, 1.0) (89.0, 0.99) (79.0, 0.98) (68.0, 0.96) (58.00000000000001, 0.93) (47.0, 0.88) (37.0, 0.8) (26.0, 0.68) (16.000000000000004, 0.49) (5.000000000000004, 0.2) (0, 0)
    };    \addplot[ solid,blue!50,forget plot] coordinates {
(100.0, 1.0) (89.0, 1.0) (79.0, 1.0) (68.0, 0.99) (58.00000000000001, 0.99) (47.0, 0.97) (37.0, 0.93) (26.0, 0.86) (16.000000000000004, 0.68) (5.000000000000004, 0.31) (0, 0)
    };    \addplot[ solid,blue!50,forget plot] coordinates {
(100.0, 1.0) (89.0, 1.0) (79.0, 1.0) (68.0, 0.99) (58.00000000000001, 0.98) (47.0, 0.95) (37.0, 0.91) (26.0, 0.81) (16.000000000000004, 0.63) (5.000000000000004, 0.27) (0, 0)
    };    \addplot[ solid,blue!50,forget plot] coordinates {
(100.0, 1.0) (89.0, 1.0) (79.0, 1.0) (68.0, 1.0) (58.00000000000001, 1.0) (47.0, 0.99) (37.0, 0.98) (26.0, 0.93) (16.000000000000004, 0.79) (5.000000000000004, 0.4) (0, 0)
    };    \addplot[ solid,blue!50,forget plot] coordinates {
(100.0, 1.0) (89.0, 0.99) (79.0, 0.98) (68.0, 0.96) (58.00000000000001, 0.93) (47.0, 0.88) (37.0, 0.81) (26.0, 0.69) (16.000000000000004, 0.5) (5.000000000000004, 0.2) (0, 0)
    };    \addplot[ solid,blue!50,forget plot] coordinates {
(100.0, 1.0) (89.0, 0.99) (79.0, 0.99) (68.0, 0.97) (58.00000000000001, 0.95) (47.0, 0.91) (37.0, 0.84) (26.0, 0.72) (16.000000000000004, 0.53) (5.000000000000004, 0.22) (0, 0)
    };    \addplot[ solid,blue!50,forget plot] coordinates {
(100.0, 1.0) (89.0, 0.99) (79.0, 0.98) (68.0, 0.96) (58.00000000000001, 0.93) (47.0, 0.88) (37.0, 0.8) (26.0, 0.68) (16.000000000000004, 0.49) (5.000000000000004, 0.2) (0, 0)
    };
    \addplot[solid,blue!50,forget plot] coordinates {
      (100,1.00) (90,0.89) (80,0.79) (70,0.68)
      (60,0.58) (50,0.47) (40,0.37) (30,0.26)
      (20,0.16) (10,0.05) (0,0.00)
    };

  \end{groupplot}
\end{tikzpicture}
    \vspace{-0.5em}
    \caption{{
    NLL vs KV retention on 13 tasks from RULER, one line per sub-task. For each sub-task we compute the (reciprocal) gain in NLL of the ground-truth answer when conditioned on compressed contexts (for each compression method). This empirical trend motivates the exponential calibration function introduced in \S\ref{method:cal_compress}}. These figures are generated without head-adaptive compression.}
    \label{fig:nll_curves}
    \vspace{-1em}
\end{figure}
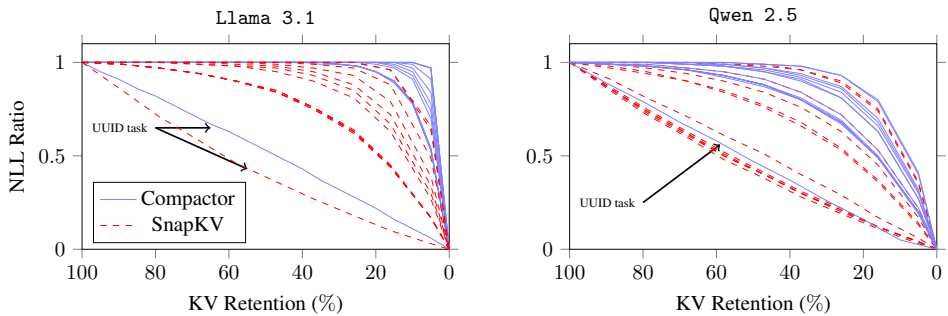

For each eviction mechanism we fit the context-calibration equation $f_{\alpha, \beta}$, (see \S\ref{method:cal_compress}) using tuples of $(r, c, y)$ generated from a subset of RULER. Figure~\ref{fig:nll_curves} shows the (mean) NLL vs KV retention curves ($r$ vs $y$) obtained by applying the \method eviction mechanism in comparison to SnapKV. \method significantly outperforms SnapKV. Note the blue lines along the diagonal of the plot corresponds to the RULER sub-task of UUID needle retrieval, a highly incompressible task; as expected, the NLL gain increases linearly with as KV retention falls. This demonstrates the ability of the proposed calibration function to recognize incompressible text.

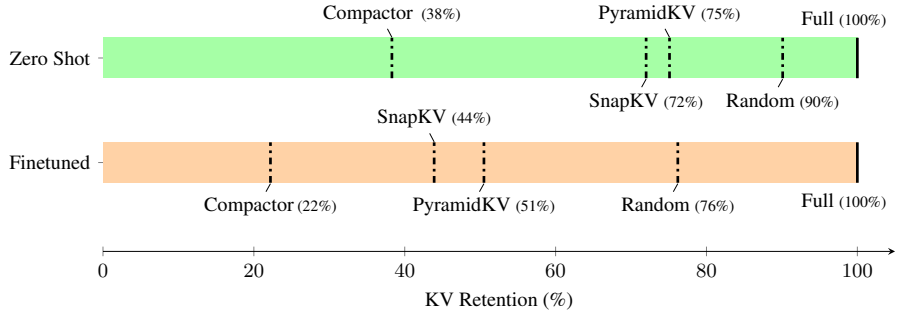
\begin{figure}[h]
    \centering
\begin{tikzpicture}[every node/.style={font=\small}, scale=0.85]
  \newcommand{\tickmarklow}[4]{%
    \draw[#2] ([yshift=-9pt]axis cs:#1,#4) -- ([yshift=9pt]axis cs:#1,#4);
    \draw[black] ([yshift=-9.4pt]axis cs:#1,#4)
               -- ([xshift=-2pt,yshift=-13pt]axis cs:#1,#4);
    \node[anchor=north,yshift=0pt] at ([yshift=-12pt]axis cs:#1,#4) {#3};
  }
  \newcommand{\tickmarkhigh}[4]{%
    \draw[#2] ([yshift=-9pt]axis cs:#1,#4) -- ([yshift=9pt]axis cs:#1,#4);
    \draw[black] ([yshift=9.2pt]axis cs:#1,#4)
               -- ([xshift=-2pt,yshift=13pt]axis cs:#1,#4);
    \node[anchor=south,yshift=0pt] at ([yshift=12pt]axis cs:#1,#4) {#3};
  }
  \begin{axis}[
      xbar, 
      width=\linewidth, height=6cm,
      xmin=0, xmax=105,
      bar width=18pt,
      symbolic y coords={Finetuned,Zero Shot},
      ytick=data,                              
      axis y line=left,
      axis x line=bottom,
      xtick={0, 20, 40, 60, 80, 100},                            
      xlabel={KV Retention (\%)},
      enlarge y limits=.50,                   
      y axis line style={opacity=0},
  ]
        
        \addplot+[fill=green!35,draw=none] coordinates {(00,Finetuned)(00,Zero Shot)};
        
        \addplot+[fill=orange!35,draw=none, forget plot] coordinates {(100,Finetuned)};
        \addplot+[fill=green!35,draw=none, ] coordinates {(100,Zero Shot)};
        \addplot+[fill=green!35,draw=none] coordinates {(00,Finetuned)(00,Zero Shot)};
        \draw[black,very thick] ([yshift=-9pt]axis cs:100,Finetuned) -- ([yshift=9pt]axis cs:100,Finetuned);
        \node[anchor=north,yshift=-1pt,xshift=-6pt] at ([yshift=-9pt]axis cs:100,Finetuned) {{Full {\scriptsize (100\%)}}};
        
        \draw[black,very thick] ([yshift=-9pt]axis cs:100,Zero Shot) -- ([yshift=9pt]axis cs:100,Zero Shot);
        \node[anchor=south,yshift=1pt,xshift=-6pt] at ([yshift=9pt]axis cs:100,Zero Shot) {{Full {\scriptsize (100\%)}}};
        
        \tickmarkhigh{32}{black,very thick, dashdotted}{\shortstack{{\method} {\scriptsize (32\%)}}}{Zero Shot}
        \tickmarkhigh{51}{black,very thick, dashdotted}{\shortstack{SnapKV {\scriptsize (51\%)}}}{Zero Shot}
        \tickmarklow{46}{black,very thick, dashdotted}{\shortstack{KeyDiff \\{\scriptsize (48\%)}}}{Zero Shot}
        \tickmarklow{90.1}{black,very thick, dashdotted}{\shortstack{Random \\{\scriptsize (90\%)}}}{Zero Shot}
        \tickmarklow{54}{black,very thick, dashdotted}{\shortstack{TOVA \\{\scriptsize (53\%)}}}{Zero Shot}

        \tickmarkhigh{19}{black,very thick, dashdotted}{{\method {\scriptsize (19\%)}}}{Finetuned}
        \tickmarklow{28}{black,very thick, dashdotted}{{KeyDiff {\scriptsize (32\%)}}}{Finetuned}
        \tickmarkhigh{40.2}{black,very thick, dashdotted}{{SnapKV {\scriptsize (40\%)}}}{Finetuned}
        \tickmarklow{76.2}{black,very thick, dashdotted}{{Random {\scriptsize (76\%)}}}{Finetuned}
        \tickmarklow{42.2}{black,very thick, dashdotted}{{TOVA {\scriptsize (42\%)}}}{Finetuned}
  \end{axis}
\end{tikzpicture}
\vspace{-.6em}
    \caption{The top bar shows the KV retention rates that each compression method induces when using context-calibrated compression on Longbench. The bottom bar shows the same when the LLM is finetuned on documents from the Longbench test set (no queries). In all cases, the performance of the compression methods is within 0.1 of full KV cache performance ($42.4\pm0.1$). Results are shown for \texttt{Llama-3.1 8B}.}
    \label{fig:longbench_calibrated_bars}
    \vspace{-1em}
\end{figure}

We next evaluate the context-calibration curves fit on RULER on Longbench (Figure~\ref{fig:longbench_calibrated_bars}).
For each test sample, we first infer the compression ratio $r$ supported by a given context while maintaining a quality budget (NLL ratio) of $\tau=0.95$ (i.e we accept a $\approx5\%$ increase in NLL relative to the uncompressed cache). We then retain $r$ fraction of the tokens according to each method's eviction procedure.

For all tested methods, the performance when applying context-calibrated compression is within $.5\%$ of uncompressed cache performance. This confirms both the utility of NLL as a proxy for downstream task performance and the efficacy of the proposed calibration function at ensuring compression does not degrade performance. Note \method  is able to achieve this performance while retaining  fewer tokens. In particular, SnapKV's is far removed from the excellent performance it demonstrates in the query-aware setting \citep{li_snapkv_2024}, where it can match baseline performance while retaining only $10\%$ of the cache.

Lastly, we finetune the model on documents from Longbench (no queries or answers), and evaluate the performance of context-calibrated compression with the finetuned model on Longbench (we evaluate and finetune with the same set of documents). We would expect that the NLL of the documents to fall, allowing for higher compression rates for the documents. Such a procedure is useful when we know \emph{a priori} which corpus will be assessed, and we desire the highest possible accuracy and compression on contexts from this corpus. For example, a question-answering LLM may be finetuned on a corpus of supporting documents, while simultaneously being presented with relevant retrieved documents from that corpus at test time. The result of this experiment is shown in Fig~\ref{fig:longbench_calibrated_bars} (``Finetuned"). As expected, we see that all models achieve higher compression rates when finetuned on the test corpus. \method still significantly outperforms competing techniques, though the gap is shortened; notably, \method without training (Zero Shot) outperforms SnapKV and TOVA even after finetuning

\subsection{Ablations}\label{sec:ablation}

\begin{wraptable}{r}{0.5\textwidth}
\vspace{-5em}
  \centering
  \caption{Ablation study on RULER of \method variants on \texttt{Llama 3.1}. Values represent the mean score across all 13 RULER tasks.}
  \vspace{1em}
  \scalebox{0.9}{\begin{tabular}{lccccc}
    \toprule
    & \multicolumn{5}{c}{KV Retention}\\
    \cmidrule(r){2-6}
    Variant & 75\% & 50\%& 25\% & 10\% \\
    \midrule
    Full & 95.6 & 94.7 & 83.1 & 64.8  \\
    \addlinespace[0.25em]
    \quad– Exact $\ell_i$   & 95.5 & 94.9 & 82.8 & 64.7  \\
    \quad– $k=64$   & 95.3 & 94.2 & 83.2 & 64.8 &  \\
    \quad– $\lambda = 0.4$   & 95.7 & 94.8 & 82.3 & 63.8   \\
    \quad– $\lambda = 0.2$   & 94.1 & 94.3 & 83.1 & 63.0   \\
    \quad– $\lambda = 0.15$   & 94.3 & 94.5 & 80.1 & 62.3   \\
    \quad– Only $\vec a$   & 93.5 & 79.0 & 61.3 & 35.7   \\
    \quad– Only $\vec o$   & 96.1 & 94.1 & 73.5 & 44.7   \\
    \bottomrule
  \end{tabular}}
    \label{tab:ablation}
    \vspace{-2em}
\end{wraptable}

Finally, we assess the necessity of both components of the scoring mechanism, the stability of the performance under choice of $\lambda$, and the result of using the exact leverage scores (Table~\ref{tab:ablation}). The addition of attention scores $\vec a$ greatly improves the performance of the eviction mechanism, especially at lower compression ratios, while the choice of $\lambda$ has minimal impact. Usage of approximate leverage scores has little impact on performance, while providing significant speed-ups. 

\section{Conclusion}

We introduce \method,  a lightweight, \emph{query-agnostic} strategy for KV cache pruning that delivers three concrete benefits over prior work: (1) higher accuracy at higher compression, across both synthetic (RULER) and real-world (LongBench) datasets; (2) \method scoring incurs minimal overhead compared to the cost of a single Flash Attention \citep{dao_flashattention_2022} pass for large contexts; and (3)  \method is training-free at inference and orthogonal to other memory-saving techniques. Finally, we are the first to introduce  \textit{context-calibrated} compression for automatic compression ratio selection, which is applicable to any compression method. We believe \method is an immediately practical drop-in for long-context LLM serving.

\bibliography{main}
\bibliographystyle{tmlr}

\appendix
\section{Proof of Theorem~\ref{thm:approx_lev_scores}}
\label{appendix:proof_thm_leverage_approximation}
\setcounter{thm}{2}
\begin{usethmcounterof}{thm:approx_lev_scores}
    Let data matrix $\K \in \R^{N \times d}$ and target dimension $k=\lceil r \cdot N \rceil$ be given. Let distortion factor $\epsilon \in (0, 1)$ and failure probability $\delta \in (0, 1)$ be given. Define $\mathbf \Phi \in \mathbb R^{d \times k}$ be a matrix whose entries are drawn i.i.d from $\mathcal N (0, \frac{1}{k})$. Compute the SVD of the sketched matrix $\mathbf {K \Phi}$ and approximate leverage scores $\tilde \ell_i$ as: 
    \[
    \textsc{svd}\left(\K\mathbf\Phi \right)= \mathbf {\tilde U} \mathbf{\tilde \Sigma} \mathbf{\tilde V}^\top \qquad \qquad \tilde \ell_i = ||\tilde U_i||_2^2 
    \]
    If $k \geq 12 \epsilon^{-2} \left(\operatorname{rank}(\K) \log (42\epsilon^{-1}) + \log(2\delta^{-1})\right)$
    Then we have that with probability $1- \delta$
    \[
    \kappa(\K)^{-1}\frac{1 - \epsilon}{(1 + \epsilon)} \ell_i \leq  \tilde \ell_i \leq  \kappa(\K) \frac{1 + \epsilon}{1 - \epsilon} \ell_i \qquad \qquad \forall i \in [N]
    \]
    where $\kappa(\K)$ is the condition number of $\K$.
\end{usethmcounterof}
\begin{proof}
    The proof of the above is straightforward. It relies on the following intermediary result from \citep{gilbert_sketched_2012}:
    \begin{cor}
        \label{cor:gilbert}
        Suppose $\mathbf \Phi$ and $k$ are chosen as above, then we have that with probability $1 - \delta$ 
        \[
        (1 - \epsilon) \leq \frac{(\sigma'_j)^2}{\sigma^2_j} \leq (1 + \epsilon)
        \]
        where $\sigma_j$ denote the singular values of $\mathbf K$ and $\sigma'_j$ denote the singular values of $\mathbf K\Phi$.
    \end{cor}
    Note that Corollary~\ref{cor:gilbert} also applies to the case of right sketching $\mathbf K$. Now consider the following SVDs 
    \[
    \mathbf K\mathbf\Phi = \tilde {\mathbf U} \tilde {\mathbf\Sigma} \tilde {\mathbf V}^\top  
    \qquad \qquad     \mathbf K = \mathbf U \mathbf \Sigma \mathbf V^\top  
    \]
    Because $\Phi$ is a subspace embedding, we have the following relationships between the norms of each row $K_i$ and $(K \Phi)_i$
    \[
    (1 - \epsilon)||K_i||^2_2 \leq ||(\mathbf K \mathbf \Phi)_i||_2^2 \leq (1 + \epsilon)||K_i||^2_2
    \]
    Because $\mathbf V$ is a unitary transformation, we further have that 
    \[
    (1 - \epsilon)||U_i\mathbf\Sigma||^2_2 \leq ||\tilde U_i \tilde {\mathbf \Sigma}||_2^2 \leq (1 + \epsilon)||U_i \mathbf\Sigma||^2_2
    \]
    We can define $m = \min_{i} \mathbf\Sigma_i^2$ and $M = \min_{i} \mathbf\Sigma_i^2 $ and $m' = \min_{i} \tilde {\mathbf{\Sigma}}_i^2$ and $M' = \min_{i} \tilde {\mathbf \Sigma_i}^2 $. It follows that $m' \leq (1 - \epsilon) m $ and $M' \leq (1 + \epsilon) M$. This implies the following bounds,
    \begin{equation*}
            \|\tilde U_i\|^{2}\geq\frac{1}{M'}\|\tilde U_i\tilde {\mathbf \Sigma}\|^{2}\ge\frac{1-\varepsilon}{M'}\|U_i\mathbf \Sigma\|^{2}\ge \frac{1-\varepsilon}{M'}m\|U_i\|^{2}
    \end{equation*}
\[
\|\tilde U_i\|^{2}\le\frac{1}{m'}\|\tilde U_i\tilde{\mathbf \Sigma}\|^{2}\le\frac{1+\varepsilon}{m'}\|U_i\mathbf\Sigma\|^{2}\le\frac{1+\varepsilon}{m'}M\|U_i\|^{2}
\]
Denote the condition number $\kappa_\Sigma = \frac{M}{m}$, and inserting the above bounds yields the following:
\[
\frac{1-\varepsilon}{1+\varepsilon}\frac{1}{\kappa_\Sigma}\|U_i\|^{2}
\leq
\|\tilde U_i\|^{2}
\le
\frac{1+\varepsilon}{1-\varepsilon}\kappa_\Sigma\|U_i\|^{2}
\]
which immediately implies the stated bound.
\end{proof}
\section{Non-Causal Attention Matrices}
\label{appendix:non_causal_attn_mat}
{\begin{figure}[h]
  \centering
  \includegraphics[width=0.49\textwidth]{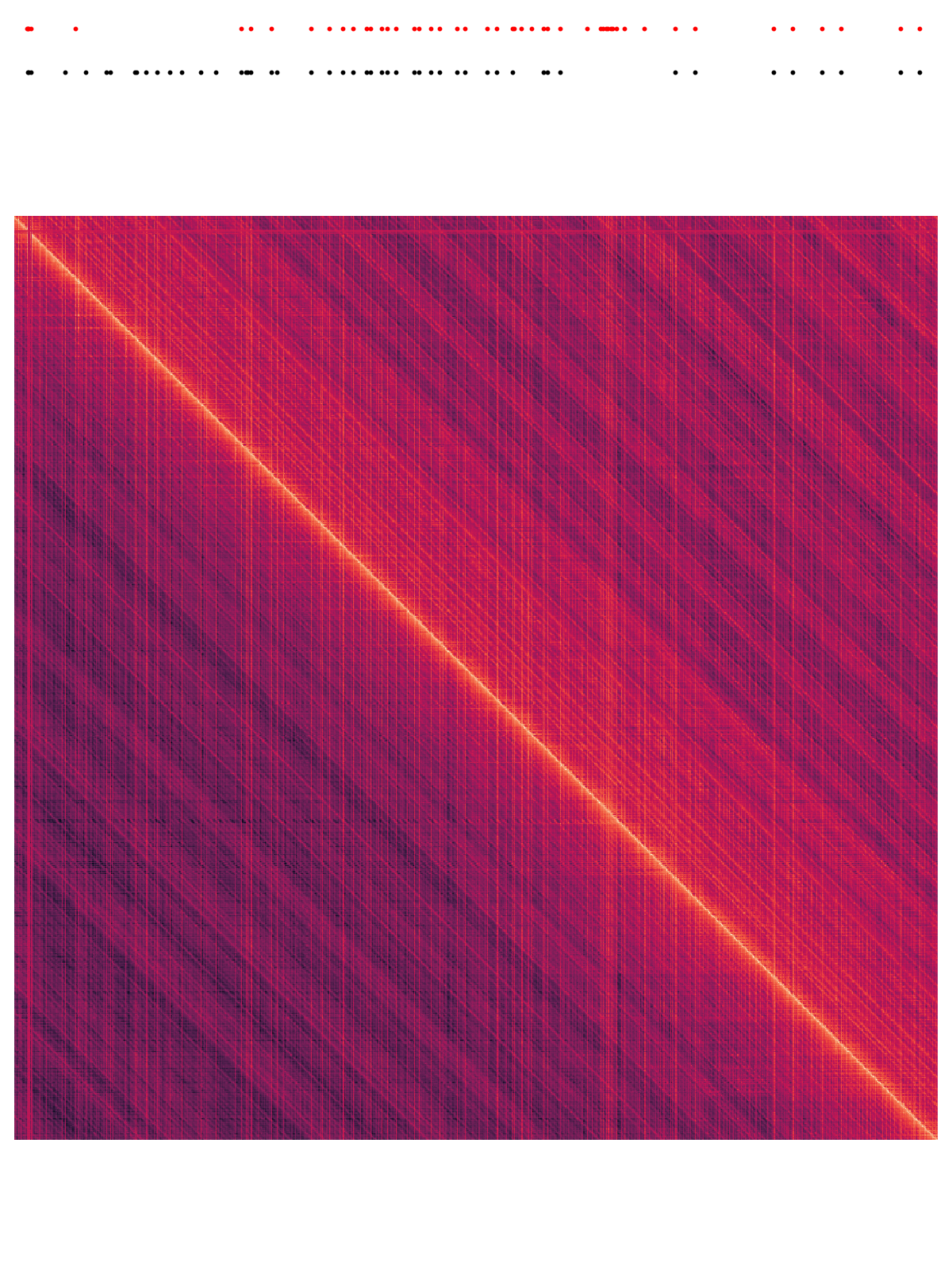}
  \includegraphics[width=0.49\textwidth]{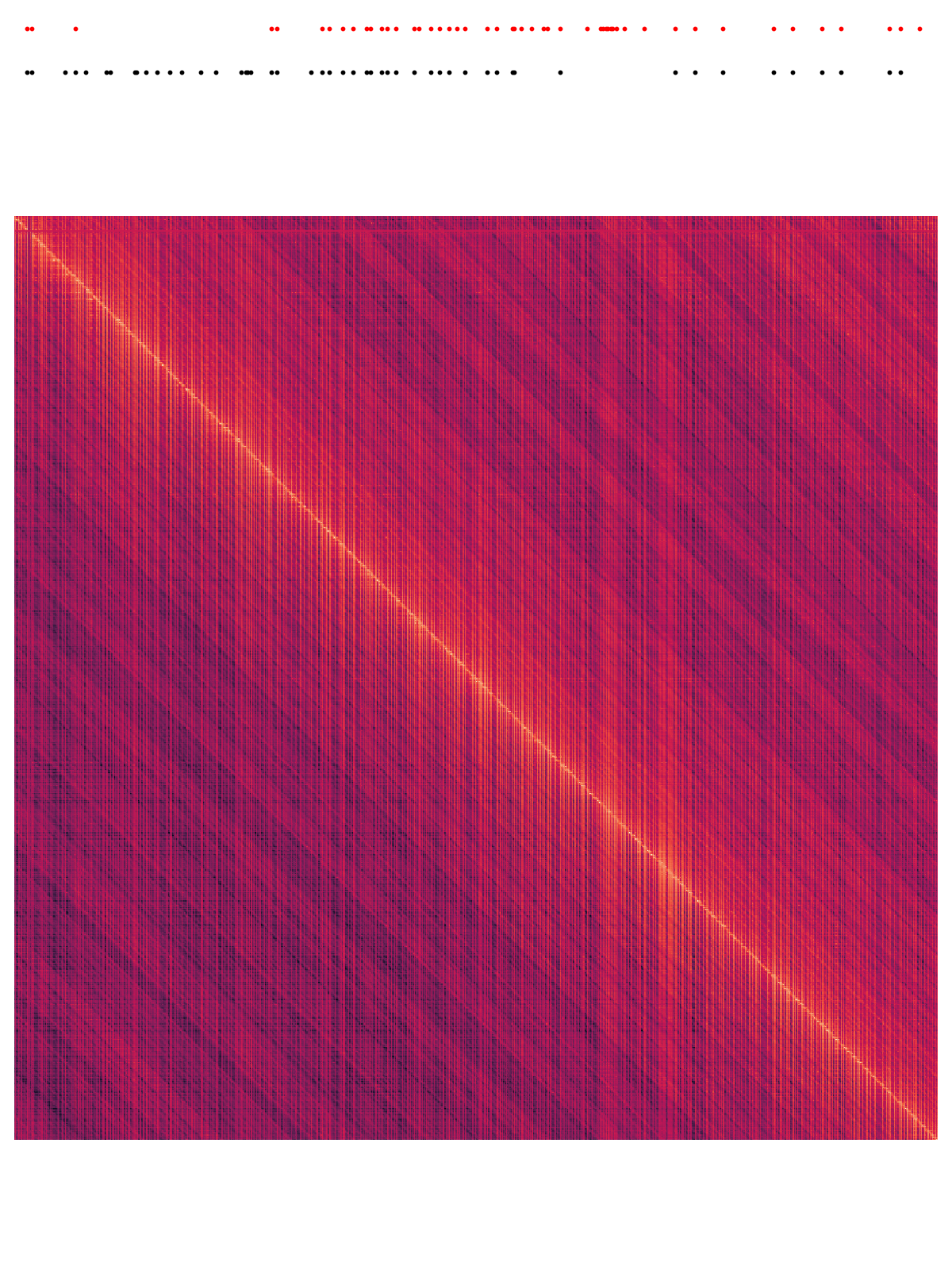}

  \caption{Example of non-causal attention matrices from head 1 and 2 in layer 16 in \texttt{Llama 3.1-8B Instruct}. Lighter colors indicate higher attention scores. Above the matrices, we first include the indices of top $5\%$ of tokens selected by using the non-causal \method $\vec a$ scores (in red, top row), and then, by SnapKV scores, (in black, bottom row)}
  \label{fig:non_causal_attn_mats_appendix}
\end{figure}}
In the main text, we used non-causal self-attention (attention without the causal mask) to obtain query-agnostic estimates of how likely any past token will be referenced by future tokens. Empirically, these matrices reveal highly structured patterns that are otherwise suppressed in the causal view. Using these patterns lets us rank tokens by the column-wise sum of their non-causal scores, producing the attention component $\mathbf a$ used by \method. One feature of the matrices is the presence of ``anchor columns"; qualitatively, at these positions, we usually find that separator tokens ($\texttt{\textbackslash n}$, $\texttt{,}$, $\texttt{...}$, $\texttt{.}$), and conjunctions ($\texttt{and}$, $\texttt{but}$) or system prefixes attract disproportionately high attention from many positions, producing bright vertical stripes. We also observe strong locality windows in some heads; even without the mask, attention mass is biased towards a narrow band around the main diagonal, demonstrating the head’s strong local inductive bias.

In Figure~\ref{fig:non_causal_attn_mats_appendix} we show an example of additional heads from Layer 16, along with the tokens selected by \method and SnapKV for these particular heads. This example is taken from the "passage retrieval" task on Longbench, in which the model is given 30 paragraphs from Wikipedia, along with an abstract, and must determine which paragraph belongs to the article the abstract is from. In both head 1 and head 2, we observe similar selections in most areas under \method and SnapKV. However, in the \method selection (upper dots, in red) we observe a tightly clustered region of selected tokens ($\approx 1/3$ of the length from the end) where the paragraph corresponding to the abstract begins. The paragraph is not identified by the SnapKV mechanism (as indicated by the lack of black dots, second line, in that portion of the line) but is identified by \method. SnapKV doesn't get this question correct, while \method does. This qualitatively demonstrates the improvement that non-causal attention scores can provide for token selection.

\section{Subroutines for Computing Approximate Leverage Scores}
\label{appendix:other_leverage_algorithms}

Throughout the main text we compute an SVD of ${\hat {\mathbf K}} 
 ={\mathbf K \Phi}\in\R^{N\times k}$ to obtain approximate leverage scores.  Two potentially cheaper, but  less numerically robust alternatives are worth documenting: QR decomposition and the eigendecomposition of the Gram matrix.  Both swap a $k\times k$ singular–value decomposition for an alternative operation and therefore (might) admit faster implementations on hardware with optimized primitives; however, they differ in their numerical stability. In this section we assess whether differences in the numerical stability of the methods impact the downstream performance of the methods. We first outline the methods in further detail. 
\paragraph{QR} Given sketched keys ${\hat {\mathbf K}}$, compute the reduced QR factorization ${\hat {\mathbf K}}= \mathbf Q\mathbf R$ where $\mathbf Q\in\R^{N\times k}$ has orthonormal columns and $\mathbf R\in\R^{k\times k}$ is upper-triangular.  The leverage score of the $i$-th token is then
\[
\tilde{\ell}_i \;=\;\bigl\lVert \mathbf Q_{i,:}\bigr\rVert_2^{\,2},
\quad i=1,\dots,N.
\]
While QR is backward-stable, the orthogonality of $\mathbf Q$ degrades when $\kappa({\hat {\mathbf K}})$ is large. Re-orthogonalization alleviates this but reduces the potential speed-up. 

\paragraph{Gram–Matrix Eigendecomposition}
Form the $k\times k$ Gram matrix
$\mathbf G = {\hat {\mathbf K}}^{\!\top}{\hat {\mathbf K}}$ and
compute its eigendecomposition
$\mathbf G = \mathbf V\mathbf\Lambda\mathbf V^{\!\top}$,
$\mathbf\Lambda=\operatorname{diag}(\lambda_1,\dots,\lambda_k)$.
Then recover an orthonormal left basis via
\[
\widetilde{\mathbf U}=\widehat{\mathbf K}\mathbf V\mathbf\Lambda^{-1/2}
\]
and set
$\tilde{\ell}_i = \lVert \widetilde{\mathbf U}_{i,:}\rVert_2^{\,2}$. The eigendecomposition is the most sensitive of the three routes. Small singular values of ${\hat {\mathbf K}}$ are squared and may fall
below machine precision, inflating the corresponding
$\mathbf\Lambda^{-1/2}$ entries and corrupting
$\tilde{\mathbf U}$.  Clamping $\lambda_j\leftarrow\max(\lambda_j,\varepsilon)$ with
$\varepsilon\!\approx\!10^{-6}$ largely mitigates the issue.

\begin{table}[h]
  \centering
  \caption{\method RULER scores computed using SVD, QR, and Eigendecomposition on \texttt{Llama 3.1 8B}. Values represent the mean score across all 13 RULER tasks.}
  \scalebox{0.9}{\begin{tabular}{lcccc}
    \toprule
    & \multicolumn{4}{c}{KV Retention}\\
    \cmidrule(r){2-5}
    Variant & 75\% & 50\%& 25\% & 10\%  \\
    \midrule
    SVD &  95.6 & 94.7 & 83.1 & 64.8\\
    QR &  95.5 & 94.4 & 83.1 & 64.9\\
    Eig.  & 95.6 & 94.7 & 83.0 & 64.7\\
    \bottomrule
  \end{tabular}}
    \label{tab:leverage_methods}
\end{table}
Empirically we observe minimal differences between the performance of the SVD and QR variants of \method on RULER. As expected, the reduced numerical stability of the eigendecomposition routine results in slightly reduced performance, especially at lower retention rates; however regardless of computation subroutine, \method still significantly improves over competing methods (see Table~\ref{fig:ruler_graph}). We conclude \method is numerically relatively robust to the particulars of the implementation, further confirming the utility of the approximate leverage scores.

\section{Subsampled Randomized Hadamard Transform}
\label{appendix:srht}
The {subsampled randomized Hadamard transform} (SRHT) provides an alternative right–sketching matrix
$\mathbf\Phi\in \R^{d\times k}$ that combines three operations:
\[
\mathbf\Phi
\;=\;
\sqrt{\tfrac{d}{k}}
\;\mathbf R\,\mathbf H\,\mathbf D,
\qquad
\mathbf H\in\{-1,1\}^{d\times d}.
\]
where $\mathbf D$ is a diagonal matrix whose entries are i.i.d. Rademacher ($\pm1$) variables, $\mathbf H$ is the Walsh–Hadamard matrix (which can be applied with the fast Hadamard transform in time $O(d\log d)$), and $\mathbf R$ selects $k$ columns uniformly {without} replacement. The above construction of $\mathbf \Phi$ satisfies the same bounds in Theorem~\ref{thm:approx_lev_scores} as the Gaussian sketch \citep{ailon_fast_2009}. The key advantage of utilizing a SRHT is that it can be computed in-place on the KV cache matrices, which  can reduce peak memory usage. Empirically we observe that substituting the SRHT for the Gaussian right sketch (all else equal) yields negligible differences in performance on RULER with \texttt{Llama-3.1-8B}. In conjunction with Appendix~\ref{appendix:other_leverage_algorithms}, these results further demonstrate that \method is relatively robust to implementation details.

\begin{table}[h]
  \centering
  \caption{\method RULER scores computed using Gaussian sketch and SRHT on \texttt{Llama 3.1 8B}. Values represent the mean score across all 13 RULER tasks.}
  \vspace{0.2em}
  \scalebox{0.9}{\begin{tabular}{lcccc}
    \toprule
    & \multicolumn{4}{c}{KV Retention}\\
    \cmidrule(r){2-5}
    Variant & 75\% & 50\%& 25\% & 10\%  \\
    \midrule
    Gaussian & 95.6 & 94.7 & 83.1 & 64.8\\
    SRHT & 95.4 & 94.3 & 84.1 & 64.0\\
    \bottomrule
  \end{tabular}}
    \label{tab:leverage_methods}
\end{table}
\end{document}